\documentclass{article}

\usepackage[ngerman, american]{babel}
\usepackage{cite}
\usepackage[pdftex]{graphicx}
\usepackage{amsmath}
\usepackage{amssymb}
\usepackage{footnote}
\usepackage{algorithmic}
\usepackage{array}
\usepackage{stfloats}
\usepackage[hyphens]{url}
\expandafter\def\expandafter\UrlBreaks\expandafter{\UrlBreaks
  \do\a\do\b\do\c\do\d\do\e\do\f\do\g\do\h\do\i\do\j%
  \do\k\do\l\do\m\do\n\do\o\do\p\do\q\do\r\do\s\do\t%
  \do\u\do\v\do\w\do\x\do\y\do\z\do\A\do\B\do\C\do\D%
  \do\E\do\F\do\G\do\H\do\I\do\J\do\K\do\L\do\M\do\N%
  \do\O\do\P\do\Q\do\R\do\S\do\T\do\U\do\V\do\W\do\X%
  \do\Y\do\Z}
\usepackage{hyperref}
\usepackage{booktabs}
\usepackage{rotating}
\usepackage{caption}
\usepackage{pbox}

\def\x{{\mathbf x}}
\def\L{{\cal L}}

\newcommand{\etal}{\textit{et al.\,}}
\newcommand{\eg}{\textit{e.g.}}
\newcommand{\ie}{\textit{i.e.}}

\usepackage{cleveref}
\crefname{appendix}{App.\negthinspace\,}{App.\negthinspace\,}
\crefname{chapter}{Chap.\negthinspace\,}{Chap.\negthinspace\,}
\crefname{equation}{Eq.\negthinspace\,}{Eq.\negthinspace\,}
\crefname{algorithm}{Alg.\negthinspace\,}{Alg.\negthinspace\,}
\crefname{section}{Sec.\negthinspace\,}{Sec.\negthinspace\,}
\crefname{subsection}{Sec.\negthinspace\,}{Sec.\negthinspace\,}
\crefname{subsubsection}{Sec.\negthinspace\,}{Sec.\negthinspace\,}
\crefname{figure}{Fig.\negthinspace\,}{Fig.\negthinspace\,}
\crefname{table}{Tab.\negthinspace\,}{Tab.\negthinspace\,}
\crefname{subfigure}{Fig.\negthinspace\,}{Fig.\negthinspace\,}
\crefname{subsubfigure}{Fig.\negthinspace\,}{Fig.\negthinspace\,}
\crefname{lstlisting}{Lst.\negthinspace\,}{Lst.\negthinspace\,}

\title{Fuzzy-based Propagation of Prior Knowledge to Improve Large-Scale Image Analysis Pipelines}

\begin{document}
%
\author{Johannes Stegmaier$^\ast$ \qquad Ralf~Mikut\thanks{Institute for Applied Computer Science, Karlsruhe Institute of Technology, Karlsruhe, Germany}}
\date{}

\maketitle

\begin{abstract}
Many automatically analyzable scientific questions are well-posed and offer a variety of information about the expected outcome \textit{a priori}. Although often being neglected, this prior knowledge can be systematically exploited to make automated analysis operations sensitive to a desired phenomenon or to evaluate extracted content with respect to this prior knowledge. For instance, the performance of processing operators can be greatly enhanced by a more focused detection strategy and the direct information about the ambiguity inherent in the extracted data. We present a new concept for the estimation and propagation of uncertainty involved in image analysis operators. This allows using simple processing operators that are suitable for analyzing large-scale spatiotemporal (3D+t) microscopy images without compromising the result quality. On the foundation of fuzzy set theory, we transform available prior knowledge into a mathematical representation and extensively use it enhance the result quality of various processing operators. All presented concepts are illustrated on a typical bioimage analysis pipeline comprised of seed point detection, segmentation, multiview fusion and tracking. Furthermore, the functionality of the proposed approach is validated on a comprehensive simulated 3D+t benchmark data set that mimics embryonic development and on large-scale light-sheet microscopy data of a zebrafish embryo. The general concept introduced in this contribution represents a new approach to efficiently exploit prior knowledge to improve the result quality of image analysis pipelines. Especially, the automated analysis of terabyte-scale microscopy data will benefit from sophisticated and efficient algorithms that enable a quantitative and fast readout. The generality of the concept, however, makes it also applicable to practically any other field with processing strategies that are arranged as linear pipelines.
\end{abstract}
%
%

\section{Background}
\label{sec:UncertaintyFramework}
Available prior knowledge is often not sufficiently considered by automatic processing pipelines and a great amount potentially useful extra information remains unused. Particularly in the domain of image processing and image analysis, the visual analysis of acquired image data offers a large repository of usable \textit{a priori} information that can often easily be verbalized by experts of the respective application field. Examples of the successful incorporation of prior knowledge are, \eg, the approaches described in \cite{Al-Kofahi10, Bourgine10} that make use of information about the expected object number as well as their associated physical size in order to adjust and improve seed point detection algorithms. Analogously, properties like size, shape, geometry, intensity distributions and the like can be used to improve the algorithmic performance of image segmentation algorithms \cite{Fernandez10, Lou11, Khan14a}. Such prior knowledge is often embedded into the algorithm via shape penalization terms that are appended to the energy functional of graph-cut \cite{Lou11, Vu08} or a level-set segmentation \cite{Leventon00} or by generalized Hough transforms that can detect arbitrary shapes \cite{Ballard81}. Object properties like size, shape and movement dynamics can also be used to formulate efficient correction heuristics for object tracking algorithms \cite{Bao06, Schiegg13, Amat14}.

A great but often underestimated potential for algorithmic improvements lies in the estimation of uncertainties of the automatically produced results and should ideally be considered by subsequent processing steps \cite{Santo04}. On the pixel level, this uncertainty can be used to assess the information quality of a single pixel due to sensor imperfections or temperature dependence \cite{Santo04, Maji11}. Furthermore, the localization uncertainty of geometric features such as corners, centroids, edges and lines in images has been assessed in \cite{Pal01, Chen09, Chen10, Anchini07}. An approach to evaluate the quality of image registration algorithms was presented in \cite{Kybic10}. Besides many applications from the field of quality quantification, a part of research focuses on uncertainty quantification in areas such as face recognition and other biometric technologies \cite{Callejas06, Betta11, Betta12}, the tracking of shapes in ultrasound images \cite{Zhou05} or to evaluate the impact of noisy measurements on the validity of diagnosis results \cite{Mencattini10}. An uncertainty formulation based on fuzzy set theory has been employed to perform pixel- or object-based classification tasks \cite{Boskovitz02, Tizhoosh05, Radojevic15}. A further possibility to exploit the uncertainty information is to optimize parameter values of a respective operator in a feedback fashion such that the outcome minimizes a previously defined optimization criterion as demonstrated in \cite{Khan13smps, Khan13Krusebook}. Another example is the improvement of a graph-based watershed implementation, where uncertainties are used to assess the influence of individual edges on the final segmentation outcome \cite{Straehle12}. 

Hitherto, however, a uniform approach to systematically transform, embed and use the available prior knowledge to improve both existing and new algorithms is missing. Although the sequential arrangement of processing operators is a broadly used concept in image analysis, results propagated through the pipelines are mostly not assessed by the individual pipeline components with respect to their result quality. Thus, errors made in early processing steps tend to accumulate and may negatively affect the final result quality. Additionally, many existing methods for processing tasks such as seed point detection, segmentation and tracking are often not directly applicable to large-scale 3D+t data sets due to enormous memory or computation time demands.

Throughout the present contribution, uncertainty is considered as the imperfect knowledge about the validity of a piece of extracted information produced by the respective image analysis operators and is used to derive efficient improvement heuristics to enhance image analysis pipelines \cite{Bouchon00, Stegmaier12a}. We present and apply a new concept for the estimation and propagation of uncertainty involved in image analysis operators that improves the result quality awareness of each processing operator and can be used to trigger adapted processing strategies. On the foundation of our previous work \cite{Stegmaier12a}, we use fuzzy set theory to transform available prior knowledge into a mathematical representation and extensively use it to enhance the performance of image analysis operators by data filtering, uncertainty propagation and explicit exploitation of information uncertainty for result improvements. In particular, we extend an exemplary image analysis pipeline comprised of seed point detection, segmentation, multiview fusion and tracking with uncertainty handling. After introducing the general concept, we demonstrate how simple processing operators can be extended with uncertainty handling to improve large-scale analyses of 3D+t microscopy images. All methods are quantitatively validated on a comprehensive simulated validation benchmark data set that mimics embryonic development and is inspired by epiboly movements of zebrafish embryos. Moreover, we show qualitative results obtained with the presented framework on large-scale light-sheet microscopy data of developing zebrafish embryos.

\section{Methods}
\subsection{Uncertainty Propagation in Image Analysis Pipelines}
\subsubsection{The Image Analysis Pipeline Concept}
\label{sec:UncertaintyPropagation:PipelineConcept}
\begin{figure}[htbp]
\begin{centering}
\includegraphics[width=1.0\hsize]{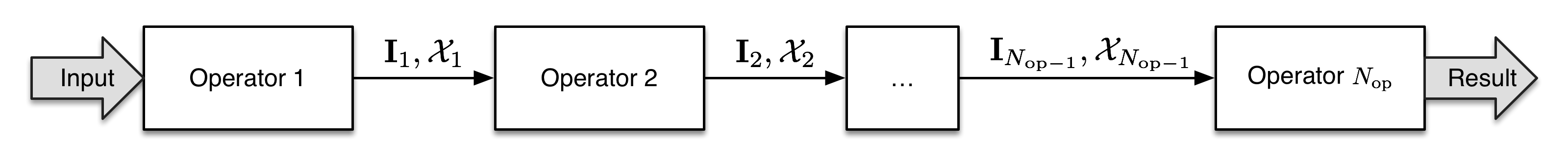}
\caption{General image analysis pipeline comprised of $N_{\text{op}}$ sequentially arranged processing operators. Each operator directly depends on the quality of the input images ($\mathbf{I_\ast}$) or features ($\mathcal{X}_\ast$) provided by its predecessor (adapted from \cite{Stegmaier12a}).}
\label{fig:OperatorPipelineReduced}
\end{centering}
\end{figure}

Most image analysis pipelines make use of multiple processing operators that are arranged as a linear processing pipeline and perform specialized tasks, such as improving or transforming the image signal, or to extract information from the images (\cref{fig:OperatorPipelineReduced}). The $N_{\text{op}}$ sequentially connected operators receive either a raw or processed version of the input image (denoted by $\mathbf{I}_i$), extracted features (denoted by $\mathcal{X}_i$) or both from their preceding processing operator with $i \in \{1, ..., N_\text{op} \}$ being the ID of the operator.
The output set $\mathcal{X}_i$ of processing operator $i$ is an ${(N_i \times N_{\text{f},i})}$ matrix with $N_i$ data tuples and $N_{\text{f},i}$ features. For processing operators without any feature output, $\mathcal{X}_i$ is an empty matrix and only the processed image is passed to the next operator.

\subsubsection{Identification of Suitable Prior Knowledge}
\label{sec:PriorKnowledgeIdentification}
Prior knowledge can be obtained from literature, expert knowledge, experimental evidence or knowledge databases. Visual analysis of acquired data often allows experts to easily identify recurring patterns, intensity properties or the appearance of objects contained in the images that can be described in natural language. An exemplary overview of such prior information derived from microscopy images is summarized in \cref{tab:PriorKnowledgeTable}. 
\begin{table*}[htbp]
\caption{Prior knowledge for 3D+t image analysis and exemplary natural language expressions.}
\begin{tabular}{m{0.12\textwidth}m{0.55\textwidth}m{0.26\textwidth}}
\toprule
\textbf{Source} & \textbf{Description} & \textbf{Example} \\
\midrule
Image Acquisition & Acquisition-specific prior knowledge such as illumination conditions, detection path, image resolution, physical spacing of voxels, high-quality image regions, point spread function (PSF) or the detection path. & Image quality decreases from XY to YZ. \\
\midrule
Intensity & Signal-dependent information like the intensity range, time-variant characteristics of objects (\eg, photobleaching in fluorescence microscopy), signal-to-noise ratio and global statistical properties of the image intensity values. & Valid objects are brighter than XY. \\
\midrule
Localization & Positional information of the objects or object properties in absolute image coordinates. Furthermore, localization of extracted properties or objects relative to each other can be used to define neighborhood relations. & Object type XY only appears close to location YZ. \\
\midrule
Spatial Extent & Object properties such as size, volume, principal components, convex hull extents or bounding volumes. & Object type X is larger than Y but smaller than Z. \\
\midrule
Geometry & Geometrical properties like dimensionality, symmetry, shape, proportions and relative localization of features within an object. & Object type XY has a line-like shape with a central symmetry axis. \\
\midrule
Morphology & Combination of intensity-based and geometrical properties, \eg, to link information about patterning, texture, structure and color to geometrical properties such as shape and symmetry. & Object XY is spherical, bright and has a textured surface. \\
\midrule
Object Interaction & Characterization of between-object properties like clustering, adhesion, repulsion, division or regional density changes. & Object behavior XY rather appears in dense regions. \\
\midrule
Spatio-Temporal Coherence & Dynamically changing quantities such as object growth, movement direction, speed, object appearance and disappearance. & Object moves maximally XY pixels between two subsequent frames. \\
\bottomrule
\end{tabular}
\label{tab:PriorKnowledgeTable}
\end{table*}
The prior information are listed in bottom-up order, \ie, from the acquisition stage over the content of a single image through to the spatio-temporal comparison between time series of images.  Naturally, the listing is not exhaustive and the suitable features have to be carefully selected to match the underlying image material and analysis problem. In the following sections, the presented natural language expressions will be used to transform the prior knowledge of different sources to a consistent mathematical representation using the concept of fuzzy sets.

\subsubsection{Quantifying Prior Knowledge using Fuzzy Set Membership Functions}
\label{sec:FuzzySetMembershipFunctions}
\begin{figure}[htbp]
	\begin{centering}
\includegraphics[width=1.0\hsize]{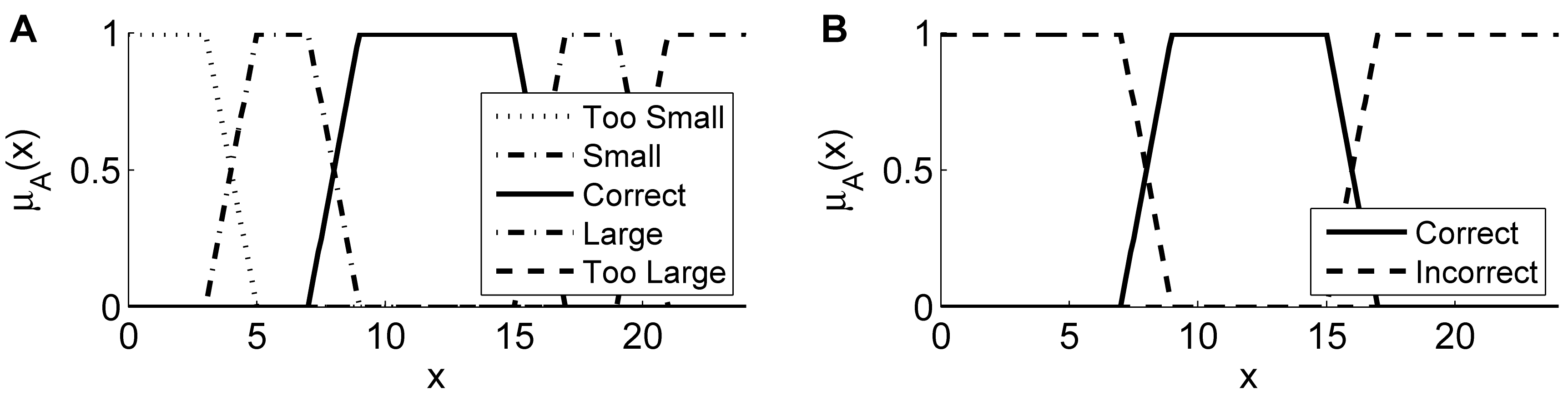}
\caption{Different possibilities to partition the input space of a feature $x$ using trapezoidal membership functions. In (A), each of the linguistic terms has a separate fuzzy set and (B) shows a reduced version with only two fuzzy sets that correspond to the desired class and its complement. In (B), different possibilities to summarize the correct objects arise. Besides restricting the class to the correct set as done in (B), the correct fuzzy set could be extended by the potentially useful classes (\textit{Small} and \textit{Large}). However, the appropriate formulation has to be chosen application dependent.}
	\label{fig:TrapezoidalMembershipFunctionComparison}
	\end{centering}
\end{figure}

To transform the prior knowledge presented in \cref{tab:PriorKnowledgeTable} into a mathematical representation, we make use of fuzzy sets that have been introduced by Zadeh in 1965 \cite{Zadeh65}. Analogous to the characteristic function of a classical set, a fuzzy set $\mathcal{A}$ can be defined by its associated membership function (MBF) $\mu_{\mathcal{A}}:\mathcal{X}\mapsto [0,1]$ that maps each element of a universe of discourse $\mathcal{X}$ to a value in the range $[0,1]$ \cite{Zadeh65}. This assigned value in turn directly reflects the fuzzy set membership degree (FSMD) of the respective element to the fuzzy set $\mathcal{A}$. The special cases $\mu_{\mathcal{A}}(x)=1$ and $\mu_{\mathcal{A}}(x)=0$ indicate that $x$ is fully included or not part of the fuzzy set $\mathcal{A}$, respectively \cite{Bede13}. The most common membership functions used in practice are trapezoidal membership functions, which can be parameterized to model singletons, triangular and rectangular MBFs. A trapezoidal membership function can be formulated as
\begin{gather}
	\mu_\mathcal{A}(x,\pmb{\theta}) = \max\left( \min\left( \frac{x-a}{b-a}, 1, \frac{d-x}{d-c} \right), 0\right),
	\label{eq:TrapezoidalMBF}
\end{gather}
with the parameter vector $\pmb{\theta}=\left(a,b,c,d\right)^\top$ that is used to control the start and end points of the respective transition regions. Here, we make use of a standard partition, \ie, maximally two neighboring fuzzy sets overlap and have non-zero membership values for a certain value of $x$ and the respective membership degrees for any input value of $x$ sum up to $1$ (\cref{fig:TrapezoidalMembershipFunctionComparison}).

Considering, for instance, the detection of a specific kind of object it might make sense to use the size feature of the object as an indicator of its appropriateness. As described in \cite{Stegmaier12}, the linguistic terms could in this case for example be determined by five possible outcomes, where the extracted feature ...
\begin{enumerate}
	\item ... perfectly matches the expected value (\textit{Correct}).
	\item ... is smaller than expected but might be useful (\textit{Small}).
	\item ... is larger than expected but might contain useful information (\textit{Large}).
	\item ... is too small and not useful (\textit{Too Small}, \eg, noise or artifacts).
	\item ... is too large and not useful (\textit{Too Large}, \eg, segments in background regions).
\end{enumerate}
Available prior knowledge can be used to determine the parameterization of the associated fuzzy sets and an exemplary standard partition is shown in \cref{fig:TrapezoidalMembershipFunctionComparison}A. If only one outcome of the operators is of importance (\eg, Case 1 in the above-mentioned example), it is possible to use only one linguistic term and to aggregate all other cases by its complement (\cref{fig:TrapezoidalMembershipFunctionComparison}B). We use $\mu_{\mathcal{A}_{ifl}}: \mathbb{R} \rightarrow [0,1]$ to denote the fuzzy set membership function for image analysis operator $i$, feature $f \in \lbrace 1,...,N_{\text{f},i} \rbrace$ and linguistic term $l \in \lbrace 1,...,N_\text{l} \rbrace$. Thus, the $n$-th data tuple produced by operator $i$ obtains the FSMD value $\mu_{\mathcal{A}_{ifl}}(x_i[n,f])$ to the fuzzy set $\mathcal{A}_{ifl}$ for each feature $f$ and each linguistic term $l$.

\subsubsection{Combination of Fuzzy Set Membership Functions}
\label{sec:MembershipFunctionCombination}
Fuzzy set membership degree values of multiple features that characterize a linguistic term (\eg, if an object of interest is bright and elongated at the same time) can be combined using a fuzzy pendant to a logical conjunction \cite{Bouchon00}. A conjunction of $N_{\text{f},i}$ fuzzy set membership functions for linguistic term $l$ can be defined using the minimum operator:
\begin{gather}
	\mu_{\text{lc},\mathcal{A}_{il}}(\mathbf{x}_{i}[n]) = \min_{f=1,...N_{\text{f},i}} \left( \mu_{A_{ifl}}(x_{i}[n,f]) \right).
	\label{eq:MembershipFunctionCombinationMin}
\end{gather}
Features that should not contribute to the combined fuzzy set membership function can be disabled by setting the corresponding MBFs to the constant value $1$ (identity element of the conjunction) and the complement of the combined linguistic term is simply given by $1-\mu_{\text{lc},\mathcal{A}_{il}}(\mathbf{x}_{i}[n])$ as illustrated in \cref{fig:TrapezoidalMembershipFunctionComparison}B. Compared to the multiplication-based conjunction used in \cite{Stegmaier12a}, the minimum-based formulation in \cref{eq:MembershipFunctionCombinationMin} is more informative, as a non-zero value directly represents the lowest FSMD value of the considered fuzzy sets.

\subsubsection{Uncertainty Propagation in Image Analysis Pipelines}
\label{sec:UncertaintyPropagation}
We use the FSMD values associated with each data tuple to perform a feed-forward propagation of the reliability of extracted data to downstream operators. For each data vector that is produced by operator $i$, we calculate the degree of membership to the respective fuzzy sets and append it to the feature output $\mathcal{X}_i$. If only a classification into correct vs.\ incorrect objects needs to be performed (\cref{fig:TrapezoidalMembershipFunctionComparison}B) or if a linguistic term is described by a combination of different fuzzy sets (\cref{eq:MembershipFunctionCombinationMin}), a single FSMD value is appended per data tuple. The FSMD values can be used to perform object filtering, extended information propagation and to resolve ambiguities.

\paragraph{Uncertainty-based Object Rejection:}
\label{sec:ObjectRejection}
The first application of the uncertainty framework is to filter the extracted output information $\mathcal{X} _i$ produced by an operator $i$ using thresholds $\alpha_{il} \in [0,1]$. According to the FSMD values $\mu_{\text{lc},\mathcal{A}_{il}}(\mathbf{x}_i[n])$ calculated for each data tuple $\mathbf{x}_i[n] \in \mathcal{X} _i$, $\mathbf{x}_i[n]$ is only passed to the next pipeline component if $\mu_{\text{lc},\mathcal{A}_{il}}(\mathbf{x}_i[n]) \geq \alpha_{il}$ for the membership to a desired set. The reduced set which serves as input for operator $i+1$ is denoted by $\tilde{\mathcal{X}}_{i}$. To keep all extracted information, the threshold is set to $\alpha_{il}=0$. Contrary, for $\alpha_{il}=1$, no uncertain information is passed to operator $i+1$. Based on application specific criteria or object properties, this FSMD-based object rejection allows to easily filter out false positive detections as demonstrated in \cref{sec:SeedDetection} and \cref{sec:Segmentation}, respectively.

\paragraph{Extended Information Propagation to Compensate Operator Flaws:}
\label{sec:InformationPropagation}
Second, we allow operators to fall back on information of penultimate processing steps if predecessors do not deliver good results. For instance, if an operator $i$ fails to sufficiently extract information from its provided input data (\eg, missing, merged or misshapen objects), it can inform downstream operators about these flawed results. Using a second threshold $\beta_{il} \in \left[\alpha_{il}, 1\right]$ for each operator, the FSMD level below which the information of the previous steps should be additionally propagated can be controlled. More formally this means that instead of only forwarding the $\alpha_{il}$-filtered set $\tilde{\mathcal{X}}_{i} \subseteq \mathcal{X}_{i}$ produced by operator $i$ to operator $i+1$ a set $\Omega_i = \tilde{\mathcal{X}}_{i} \cup \tilde{\Omega}_{i-1}$ with $i \geq 2$ and $\Omega_1 = \tilde{\mathcal{X}}_1$ is passed through the pipeline.

$\tilde{\Omega}_{i-1}$ represents the subset of elements in $\Omega_{i-1}$ which were not successfully transferred into useful information by operator $i$, \ie, elements $\mathbf{x}_{i-1}[n] \in {\Omega}_{i-1}$ that generated output $\mathbf{x}_{i}[n] \in \tilde{\mathcal{X}} _{i}$ with $\alpha_{il} \leq \mu_{\text{lc},\mathcal{A}_{il}}(\mathbf{x}_{i}[n]) < \beta_{il}$. Such elements characterize information of operator $i-1$ that might be useful in later steps to correct flawed results of operator $i$. If $\beta_{il}=1$ all information of $i-1$ that produced an uncertain outcome is propagated to the successor $i+1$. If $\beta_{il} = \alpha_{il}$ only the information $\tilde{\mathcal{X}_i}$ produced by operator $i$ is propagated. In the current version of the framework, the respective processing operators are responsible for calculating $\Omega_i$ and to appropriately calculate the respective FSMD values.

This approach was successfully used to resolve tracking conflicts that originated from under-segmentation errors as described in \cref{sec:Tracking}.




\paragraph{Resolve Ambiguities using Propagated Uncertainty:}
\label{sec:ResolvingAmbiguities}
In addition to filtering and propagating the information of the operators within the pipeline, the provided uncertainty information can explicitly be used by the processing operators to improve their results. Depending on the degree of uncertainty of provided information, parameters or even whole processing methods can be adapted if needed. Of course, the adaptions required by a particular algorithm cannot be generalized. However, we showcase two potential applications in the next sections, namely the fusion of redundant seed points (\cref{sec:SeedDetection}) and the correction of under-segmentation errors (\cref{sec:Segmentation}). \newline

\begin{figure*}[htbp]
\includegraphics[width=1.0\hsize]{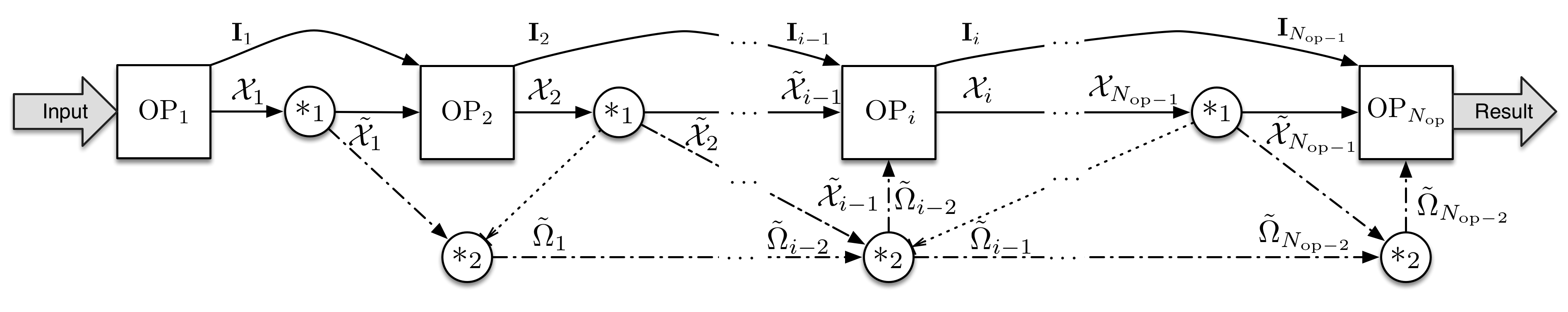}
\caption{Extended image analysis pipeline concept. Extracted output information of each operator can be filtered according to its uncertainty ($\ast_1$), operators can access information produced by penultimate predecessors ($\ast_2$) and processing operators can specifically adjust their processing behavior based on FSMD values of extracted information. Solid lines indicate the main information flow, dash-dotted lines the propagation of previously calculated results and dotted lines emphasize influence on the selection of propagated information. In addition to the flow of extracted features ($\mathcal{X}_\ast$, $\Omega_\ast$, $\mathcal{\tilde{X}}_\ast$, $\tilde{\Omega}_\ast$), the operators may pass processed images, image parts or forward the input image ($\mathbf{I}_\ast$) to the subsequent processing operator.}
	\label{fig:OperatorPipeline}
\end{figure*}

The general scheme for the proposed uncertainty propagation framework is summarized in \cref{fig:OperatorPipeline} and is applied to an exemplary image analysis pipeline in the next sections.

\subsection{Extending and Enhancing Algorithms with Uncertainty Treatment}
Based on the general concept presented in the previous section, we applied it to an exemplary image analysis pipeline comprised of seed point detection, segmentation, multiview fusion and object tracking. For each operator, FSMD values were estimated based on prior knowledge and used for algorithmic improvements where possible. We use a simulated benchmark data set that mimics 3D microscopy images containing fluorescently labeled nuclei of an artificial embryo.

The advantages of using simulated image data for validation is the possibility to have a single comprehensive data set for the validation of all pipeline components. Instead of testing each component separately on different benchmarks, this allows to uncover specific bottlenecks or error sources in the processing pipelines. Furthermore, different acquisition deficiencies such as different point-spread-functions, decreasing signal-to-noise ratios or multiview acquisition deficiencies can be simulated. The immediate availability of a reliable ground truth enables a quantitative validation without the bias observed for manually annotated benchmark data that suffers from intra- and inter-expert variability. As the simulated benchmark is close to the target application of the pipeline, namely quantitatively analyzing terabyte-scale 3D+t fluorescence microscopy images, the developed concepts and algorithms can easily be put into practice, \eg, for false positive reduction of a segmentation algorithm or for segmentation-based multiview fusion \cite{Kobitski15, Stegmaier16Diss}.

Details on the benchmark generation can be found in \cref{sec:chap4:Benchmark} and \cite{Stegmaier16a}. Abbreviations for the different algorithms are given in round brackets and a quantitative comparison of the result quality is provided in \cref{sec:Experiments}.

\subsubsection{Seed Point Detection}
\label{sec:SeedDetection}
In \cite{Stegmaier14}, a blob detection method based on the Laplacian-of-Gaussian (LoG) maximum intensity projection was used to localize fluorescently labeled cellular nuclei in 3D microscopy images. In brief, a 3D input image is filtered with differently scaled LoG filters with standard deviations $\sigma$ matching the expected object radius $r$ using the relation $\sigma= r / \sqrt{2}$. Subsequently, the 3D maximum projection of these LoG-filtered images is formed and local extrema are extracted from this projection image (LoGSM). Although, the proposed method worked well in many scenarios, it frequently missed objects that did not exhibit a strict local maximum due to an intensity plateau (\eg, elongated objects, overexposure or discretization artifacts). To get rid of this behavior, we used the $\leq$-operator instead of the $<$-operator to additionally detect non-strict local extrema (LoGNSM). However, this increased the amount of false positive detections in background regions and along elongated objects.

Detections in background regions were removed using an intensity threshold ($t_{\text{wmi}}$) applied on the mean intensity of a small window surrounding the potential detection. The remaining seed points were mostly located properly on the detected objects and remaining false positive detections largely originated from objects that were detected multiple times. To combine redundant objects to a single one, a fusion approach based on hierarchical clustering was used (LoGNSM+F). The hierarchical cluster tree was computed using Ward's minimum variance method to compute distances between clusters, \ie, the within-cluster variance was minimized to obtain equally sized clusters \cite{WardJr63}. The final clustering was obtained from the complete cluster tree using a distance-based cutoff $t_{\text{dbc}}$ that was set to the smallest expected object radius $r_{\text{min}}$, to fuse close redundant detections and to prevent fusion of neighboring objects. A single detection per object was obtained by averaging the feature vectors of all detected seeds in a cluster.

\begin{figure}[htbp]
\includegraphics[width=\hsize]{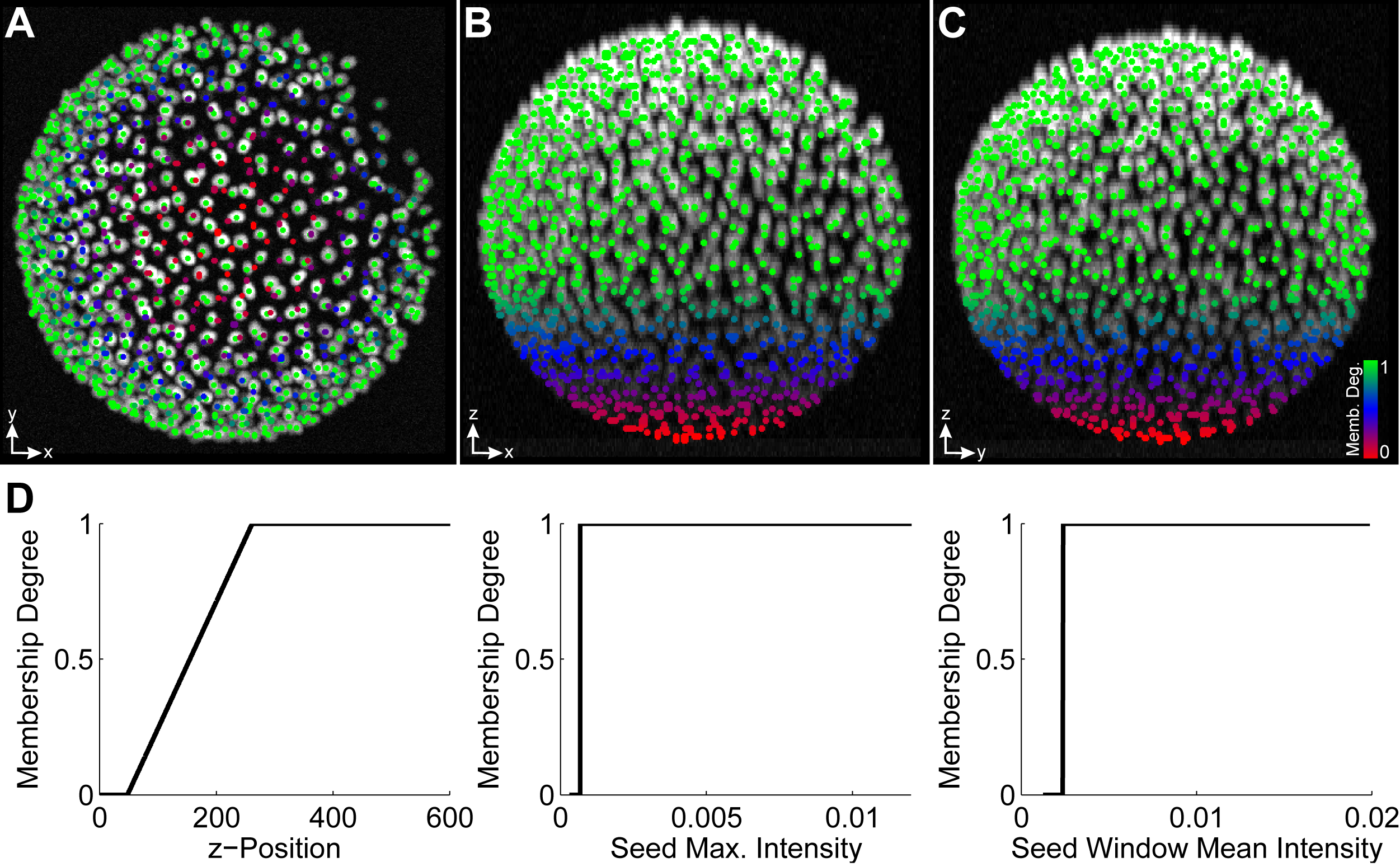}
\captionsetup{width=0.95\hsize}
\caption{Maximum intensity projection of a 3D benchmark image along the Z, Y and X axis superimposed with detected seed points (A, B, C). Seed points are colored according to their FSMD to the class of a correct detection ranging from red over blue to green for low, medium and high membership degree, respectively. The fuzzy sets used for the individual features are depicted in (D) and the \texttt{min}-operator was used as a fuzzy conjunction to obtain the final membership degree. The uncertainty gradient along the z-axis was introduced due to the signal attenuation at locations farther away from the detection objective and was used in later steps to resolve multiview fusion ambiguities.}
\label{fig:UncertaintyVisualizationAndFuzzySets}
\end{figure}

As the seed detection stage usually represents one of the first analysis steps, no preceding uncertainty information was considered. To inform down-stream operators about the expected result quality, the uncertainty of the detected seed points was estimated using the window mean intensity, the maximum seed intensity and the z-position features of the extracted objects (LoGNSM+F+U). Besides discarding obvious false positive detections in background regions (intensity-based thresholds), the fuzzy sets for the z-position were adjusted such that seed detections in low contrast regions (farther away from the detection objective) had lower membership degrees to the class of correct objects than objects in the high contrast regions (closer to the detection objective). The final fuzzy set membership degree of an object to the fuzzy set of correct objects was determined using the minimum of the obtained membership degrees and was appended as a new feature to the output matrix of the seed detection algorithm. To filter false positives the forward threshold slightly above zero to $\alpha_{11}=0.0001$ (forward threshold for Operator 1 and Linguistic Term 1). A color-coded visualization of the detected seed points and the fuzzy sets for the different features are shown in \cref{fig:UncertaintyVisualizationAndFuzzySets}.

\subsubsection{Segmentation}
\label{sec:Segmentation}
After the seed detection stage, a segmentation operator is used to extract the regions and regional properties of all detected objects from the simulated 3D image stacks, \eg, using the algorithms presented in \cite{Al-Kofahi10, Stegmaier14, Faure16}. For demonstration purposes, we further improved an algorithm based on adaptive thresholding using Otsu's method \cite{Otsu79} and a watershed-based splitting of merged objects \cite{Beare06} (OTSUWW) as described in \cite{Stegmaier14}. Therefore, propagated information from the seed detection stage and estimated FSMD values of extracted segments are used to improve the algorithmic efficiency and the segmentation quality (OTSUWW+U).

Based on the extracted statistical quantities of the benchmark images (\cref{tab:Segmentation:GroundTruthStatistics}), we derive the parameter vector $\pmb{\theta}=(a,b,c,d)^\top$ of the trapezoidal fuzzy set membership function for each considered feature using the minimum and maximum values as $a,d$ parameters, respectively. The remaining parameters $b,c$ were set to the 5\%-quantile and the 95\%-quantile. This parameterization ensured that all values smaller or larger than the maximum values obtained a membership degree of zero and that 90\% of the data range has a membership value of one assigned. Of course, this parameterization is application and data dependent and can be customized, \eg, to adjust the behavior for extrema at the lower and upper spectrum of the value range. For simplicity, the focus was put on volume and size information of the objects. In the absence of ground truth data, the transition regions for the fuzzy sets can be identified by a manual analysis of objects that deviate from the expectation at the lower and the upper feature value range, \eg, using software tools such as Fiji, ICY or Vaa3D \cite{Schindelin12, Chaumont12, Peng14}.
\begin{table}[htbp]
\caption{Statistical quantities of the benchmark data set}
\begin{center}
\resizebox{\hsize}{!}{
\begin{tabular}{lccccccc} 
\toprule
\textbf{Feat.}& \textbf{Min} & \textbf{Max} & \textbf{Mean} & \textbf{Std.} & \textbf{Med.} & \textbf{5\%\,qt.} & \textbf{95\%\,qt.} \\
\midrule
Vol. & 449 & 2016 & 993.6 & 247.2 & 990 & 617 & 1405 \\
Width & 13 & 31 & 19.9 & 2.6 & 20 & 15 & 24 \\
Height & 13 & 34 & 19.9 & 2.5 & 20 & 15 & 24 \\
Depth & 3 & 11 & 6.2 & 1.0 & 6 & 5 & 8  \\
\bottomrule
\vspace{1px}
\end{tabular}}
\end{center}
Minimum, maximum and quantile values were used to formulate fuzzy sets for each of the features. Individual fuzzy sets were combined using the minimum operator, to obtain a single membership degree value to the fuzzy set of being a valid object.
\label{tab:Segmentation:GroundTruthStatistics}
\end{table}

\begin{figure}[htbp]
\includegraphics[width=\hsize]{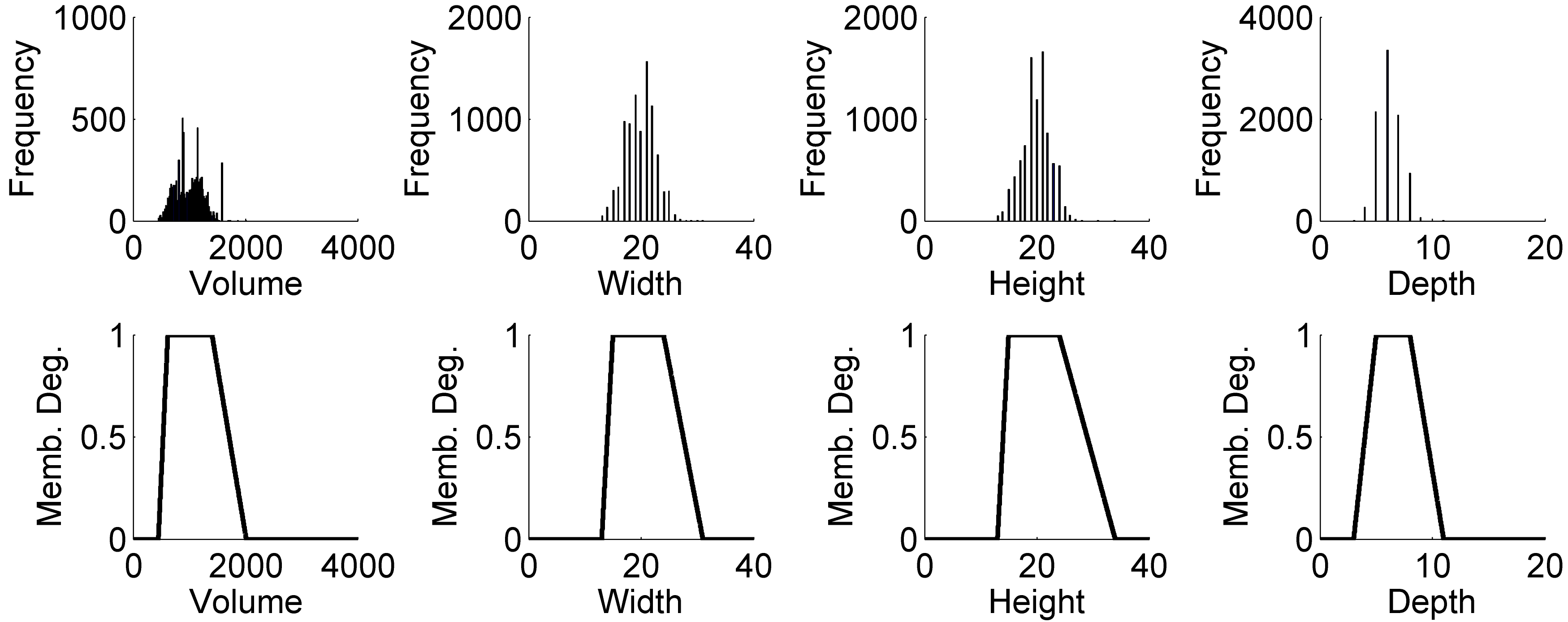}
\caption{Feature histograms (top) and fuzzy set membership functions (bottom) for volume ($\pmb{\theta}_\text{vol}=(449,617,1405,2016)^\top$), width ($\pmb{\theta}_\text{w}=(13, 15, 24, 31)^\top$), height ($\pmb{\theta}_\text{h}=(13, 15, 24, 34)^\top$) and depth ($\pmb{\theta}_\text{w}=(3, 5, 8, 11)^\top$).}
\label{fig:Segmentation:GroundTruthStatistics}
\end{figure}

As depicted in \cref{fig:Segmentation:GroundTruthStatistics}, the shapes of the fuzzy set membership functions derived from the statistical quantities resemble the respective distribution observed in the feature histograms. We used the $\min$-operator to combine the individual fuzzy set membership degrees to a single value, \ie, the combined FSMD value directly corresponded to the membership degree of the feature that deviated the most from the specified expected range. Of course, the size criteria discussed here should only be considered as an exemplary illustration. There are various other features that can potentially be used to assess and improve segmentation results, \eg, integrated intensity, edge information, local entropy, local signal-to-noise ratios (SNR), principal components, weighted centroids and many more. Furthermore, if colocalized channels are investigated, complementary information can be used to formulate more complex decision rules. In the case of a fluorescently labeled nuclei and membranes that are imaged in different channels \cite{Fernandez10, Khan14a, Stegmaier16}, rules like \textit{"each cell has exactly one nucleus"} can be formalized in the same way using the fuzzy set membership functions for a quantification of the available prior knowledge.



The identified FSMD values of segmented objects, were then used to detect under-segmentation errors produced by Otsu's method that needed be to further split to match the expected object size. This was realized in the investigated OTSUWW+U implementation using the seed points of the LoGNSM+F+U method (\cref{sec:SeedDetection}). In contrast to the previous approach (OTSUWW, \cite{Beare06, Stegmaier14}), the watershed-based splitting technique was only applied to objects that were known to be larger than expected. Using a parallelization strategy similar to the one discussed in \cite{Stegmaier14}, all segments with combined FSMD values below $\beta_{21}$ (backward threshold for Operator $2$ and Linguistic Term $1$) that corresponded to objects larger than expected (Case 3, \cref{sec:FuzzySetMembershipFunctions}) were distributed among the available CPU cores and a seeded watershed approach was used for object splitting in each of the cropped regions of the image \cite{Beare06}. This approach was much faster than directly applying the watershed algorithm on the entire image (OTSUWW), due to the uncertainty-guided, locally applied processing of erroneous objects. After splitting merged objects, the connected components of the image needed to be identified once more and the uncertainty values were re-evaluated as well, to provide the updated information to the subsequent processing operators.

To further improve the results with respect to false positive detections observed for higher noise levels, segments with combined FSMD values below $\alpha_{21}=0.1$ (forward threshold for Operator $2$ and Linguistic Term $1$), \ie, objects that were smaller than the expected object size (Case 2 and Case 4, \cref{sec:FuzzySetMembershipFunctions}) were removed from the label images. To facilitate the implementation of the uncertainty-guided segmentation, it was realized using just a single fuzzy set for the correct class of objects (Case 1, \cref{sec:FuzzySetMembershipFunctions}) to identify objects that needed further consideration. To determine if objects with low FSMD values were smaller or larger than the expectation, a comparison to the boundaries of the expected valid range was performed (trapezoidal fuzzy set parameters $b, c$). Threshold values were identified using the interactive graphical user interface presented in \cite{Stegmaier16}.

\subsubsection{Tracking}
\label{sec:Tracking}
The final step of the investigated image analysis pipeline was the tracking of all detected objects, \ie, to identify the correct correspondences of detected objects in subsequent time points of the acquired images. For illustration purposes, we used a straightforward nearest-neighbor tracking approach implemented in the open-source MATLAB toolbox Gait-CAD \cite{Stegmaier12}. Each object present in a frame was associated with the spatially closest object in the subsequent frame. This procedure was applied to every frame of the data set in order to get a complete linkage of all objects. 

In addition to tracking the results obtained from the segmentation methods introduced in the previous section, we present an alternative approach that combines a flawed segmentation with provided seeds (OTSU+NN+U). Inspired by conservation tracking methods \cite{Schiegg13}, we leave the flawed segmentation results produced by the respective algorithms unchanged and additionally provide detected seed points to the tracking algorithm instead of actually using the seed points for splitting in the image domain \cite{Stegmaier12a}. As shown in \cref{tab:Segmentation:DetectionPerformance}, the segmentation quality achieved by OTSU is in principle not reasonably usable without the watershed-based object splitting. Nevertheless, the tracking algorithm could be extended to decide which of the information was reliable and suitable for tracking and could optionally fall back to the provided seed points if the segmentation quality was insufficient. Therefore, we used the FSMD values provided by the segmentation stage. Using an empirically determined threshold of $\beta_{31} = 0.9$ (backward threshold for Operator $3$ and Linguistic Term $1$), all objects with an aggregated FSMD value lower than the threshold were not tracked with the actual segment, but with the seed points the respective segment contained. The forward threshold parameter was set to $\alpha_{31} = 0.0$ in order to report all tracking results.


\section{Results and Discussion}
\label{sec:Experiments}
The functionality of the proposed approaches was validated on simulated 3D benchmark images. The data sets contained images with different numbers of objects (\texttt{SBDE1}), different noise levels (\texttt{SBDE2}) and a set of $50$ sequential time points with $1000$ moving and interacting objects (\texttt{SBDE3}). The \texttt{SBDE3} data set additionally included a multiview simulation, \ie, at each time point, two simultaneous images from opposite direction were generated. An overview of the generated benchmark data sets is provided in \cref{tab:Appendix:BenchmarkDatasetsEmbryo} and a brief description on the data set generation is provided in \cref{sec:chap4:Benchmark} as well as \cref{fig:chap4:Benchmark:Illustration} and \cref{fig:chap4:Benchmark:TimeSeries}.

\subsection{Seed Point Detection Validation}
To validate the proposed improvements of the LoG-based seed detection algorithm, the \texttt{SBDE1} and \texttt{SBDE2} benchmark data sets were used (\cref{tab:Appendix:BenchmarkDatasetsEmbryo}) with the parameters listed in \cref{tab:chap4:SeedDetection:ParameterizationTable} and the performance measures described in \cref{sec:PerformanceAssessment}. The obtained values are summarized in \cref{tab:SeedDetection:DetectionPerformance}, whereas each entry of the table corresponds to the arithmetic mean value of the independently obtained results on the ten benchmark images of \texttt{SBDE1}.
\begin{table*}[htbp]
\caption{Quantitative assessment of the seed detection performance}
\begin{center}
\resizebox{\hsize}{!}{
\begin{tabular}{lccccccccc} 
\toprule
\textbf{Method}& \textbf{TP} & \textbf{FP} & \textbf{FN} & \textbf{Rec.} & \textbf{Prec.} & \textbf{F-Sc.} & \textbf{Dist.} & \textbf{Time (s)} & \textbf{KVox./s} \\
\midrule
LoGSM & 681.1 &	\textbf{3.3} &	202.8 &	0.77 & \textbf{1.00} & 0.87 & 1.60 & \textbf{7.23} & \textbf{7259.82} \\
LoGNSM & \textbf{813.7} & 160.0 & \textbf{70.2} & \textbf{0.91} & 0.84 & 0.87 & 1.64 & 7.66 & 6858.39 \\
LoGNSM+F & 811.7 & 4.5 & 72.2 & \textbf{0.91} & 0.99 & \textbf{0.95} & \textbf{1.59}	& 7.70 & 6819.58 \\
LoGNSM+F+U & 812.5 & 4.3 & 71.4 & \textbf{0.91} & 0.99 & \textbf{0.95} & \textbf{1.59} & 9.77 & 6170.05 \\
\bottomrule
\vspace{5px}
\end{tabular}}
\end{center}
Quantitative performance assessment of the LoG-based seed detection methods. The criteria are true positives (TP), false positives (FP), false negatives (FN), recall, precision, F-Score, the distance to the reference (Dist., smaller values are better) as well as the achieved time performance measures in seconds (smaller values are better) and voxels per second (larger values are better). All values represent the arithmetic mean of the individually processed benchmark images.
\label{tab:SeedDetection:DetectionPerformance}
\end{table*}

The quantitative analysis confirmed that the proposed extensions of LoGSM could improve the algorithmic performance by up to $9.2\%$ with respect to the F-Score. LoGSM had few false positive detections but on the other hand missed many objects due to the strict maximum detection (recall of $0.77$ and precision of $1.0$). The recall could be improved by $18.2\%$ to a value of $0.91$ by additionally allowing non-strict maxima (LoGNSM). However, this adaption concurrently raised the number of false positives and thus lowered the precision by $16.0\%$ to $0.84$, as objects with maximum plateaus were detected multiple times. These multi-detection errors could be successfully removed using the proposed fusion technique, which was reflected in an F-Score value of $0.95$ for LoGNSM+F(+U), \ie, compared to the LoGSM method, the F-Score was increased by $9.2\%$. Regarding the processing times, the additional effort for a redundant detection was almost negligible, as the non-strict maximum detection simply detected more seed points during the same iteration over the image. The seed point fusion was performed directly in the feature space and was therefore also insignificant compared to the preceding processing steps. For the feature set described here, using the uncertainty-based object rejection (LoGNSM+F+U) only slightly improved the results compared to directly fusing and filtering the data using the hard intensity threshold but increased the processing time by $35\%$. LoGNSM+F yielded almost identical results and required only $6\%$ more processing time compared to LoGSM. Nevertheless, all objects were equipped with an uncertainty value that was propagated through the pipeline and proved to be beneficial to filter, fuse and correct the extracted data in subsequent steps. In addition, it should be noted that the processing time required for the image analysis easily exceeds the fuzzy set calculations as soon as the images get larger.

\begin{figure}[htbp]
\includegraphics[width=\hsize]{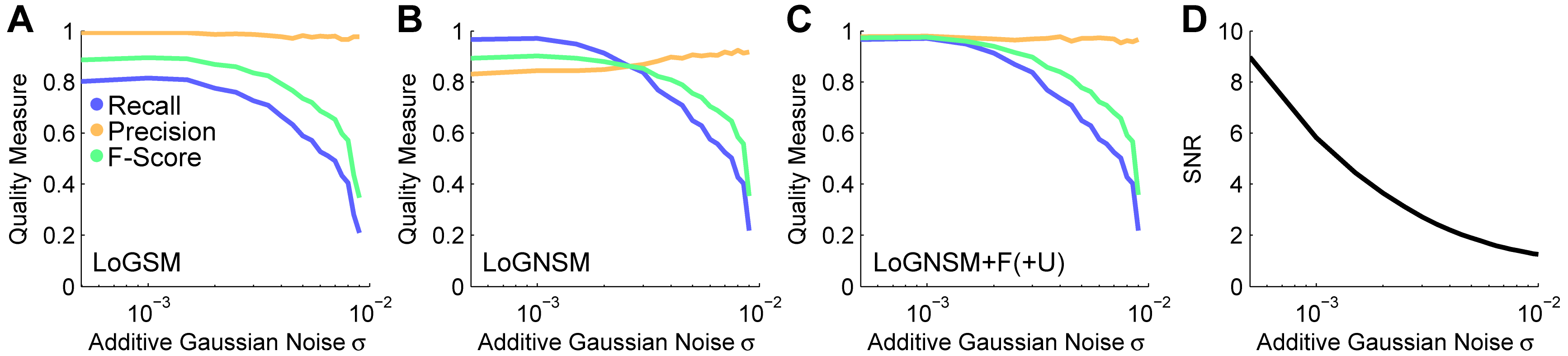}
\caption{Assessment of the seed detection performance for the different noise levels of the \texttt{SBDE2} data set. The performance measures recall, precision and F-Score are plotted versus the additive Gaussian noise level parameter $\sigma_{\text{agn}}$ for LoGSM (A), LoGNSM (B) and LoGNSM+F(+U) (C). As LoGNSM+F and LoGNSM+F+U produced identical results with respect to recall, precision and F-Score, the plots are combined to a single panel (\cref{tab:SeedDetection:DetectionPerformance}). The influence of the noise level on the signal-to-noise ratio of the images is plotted in (D).}
\label{fig:SeedDetection:NoiseLevelInfluence}
\end{figure}

Furthermore, we tested the performance of the seed detection under different image noise conditions using the \texttt{SBDE2} data set, which contained images with different settings for the additive Gaussian noise standard deviation ($\sigma_{\text{agn}} \in [0.0005, 0.01]$). Seeds from these images were extracted using the LoGSM, LoGNSM, LoGNSM+F and LoGNSM+F+U algorithms and the intensity thresholds were determined for each of the noise levels individually using the semi-automatic graphical user interface described in \cite{Stegmaier16AT}. For higher noise levels, the number of detections in background regions heavily increased and it became ambiguous to determine true positive detections in low-contrast regions. The manual threshold was therefore adjusted such that the false positive detections were minimized and only unambiguous seeds were considered. This continuous threshold adaption is also the reason for constant (\cref{fig:SeedDetection:NoiseLevelInfluence}A, C) or increasing precision (\cref{fig:SeedDetection:NoiseLevelInfluence}B), as it was easier to identify false positives rather than false negative detections in noisy image regions. Objects were robustly detected down to a signal-to-noise ratio of $5$ (\cref{fig:SeedDetection:NoiseLevelInfluence}), which was close to the visual limit of detection \cite{Murphy12} and emphasized the uncertainty-based improvements.

\subsection{Segmentation Validation}
The segmentation performance was validated using the \texttt{SBDE1} and the \texttt{SBDE2} data sets (\cref{tab:Appendix:BenchmarkDatasetsEmbryo}). In addition to the algorithms described in \cref{sec:Segmentation} (OTSU, OTSUWW, OTSUWW+U), the segmentation quality obtained by the TWANG method (see \cite{Stegmaier14} for details) was added and quantitatively compared to a TWANG version (TWANG+U) that relied on the improved seed detection operator introduced in the previous section (LoGNSM+F+U). The respective parameterization and a brief description of each algorithm is provided in \cref{tab:chap4:Segmentation:ParameterizationTable} and the validation measures are summarized in \cref{sec:PerformanceAssessment} and \cite{Coelho09}. In \cref{tab:Segmentation:DetectionPerformance}, the quantitative segmentation quality results obtained on the \texttt{SBDE1} data set are summarized. The Rand index (RI) value was almost identical for all algorithms and OTSU yielded the highest value. The enhanced adaptive threshold-based techniques yielded an 11.0\% better Jaccard index (JI) value than TWANG+U and the best normalized sum of distances (NSD) value was obtained by OTSUWW+U. Considering the results for RI, JI and NSD, the global threshold-based techniques (OTSU*) produced slightly more accurate results (0.3\%, 5.3\% and 12.0\%, respectively) for the objects they were still able to resolve compared to the best results of the TWANG-based methods. Both TWANG-based methods on the other hand produced the minimal amount of topological errors with respect to split and merged objects compared to all OTSU-based methods. The number of added objects was minimal for OTSUWW+U. TWANG+U produced slightly more added objects than TWANG but efficiently detected far more objects. Thus, the F-Score values achieved by TWANG+U were further increased by $8.4$\% compared to TWANG, \ie, TWANG+U produced the best results with the fewest topological errors (F-Score $0.9$). The low amount of split and merged nuclei for TWANG originate from the single-cell extraction strategy, rather than using a global threshold as performed in the OTSU-based methods. The relatively large amount of added objects detected by the TWANG segmentation were mostly no real false positive detections, but segments where most of the extracted region intersected with the image background instead of the actual object and are thus considered as false positives.
\begin{sidewaystable}[htbp]
\caption{Quantitative assessment of the segmentation performance}
\begin{center}
\resizebox{\hsize}{!}{
\begin{tabular}{lccccccccccccc} 
\toprule
\textbf{Method}& \textbf{RI} & \textbf{JI} & \textbf{NSD ($\times$10)} & \textbf{HM} & \textbf{Split} & \textbf{Merged} & \textbf{Added} & \textbf{Missing} & \textbf{Rec.} & \textbf{Prec.} & \textbf{F-Score} & \textbf{Time (s)} & \textbf{KVoxel/s} \\
\midrule
OTSU & \textbf{97.92} & 27.82 & 3.77 & 6.67 & 25.00 & 370.50 & 42.40 & 264.00 & 0.29 & 0.78 & 0.42 & \textbf{6.11} & \textbf{8589.46} \\
OTSUWW & 97.91 &	\textbf{27.99} &	3.29 &	6.48 &	88.60 &	96.30 &	57.90 &	276.20 &	0.58 &	0.78 &	0.67 &	42.41	&	1237.72 \\
OTSUWW+U & 97.91 & 27.98 & \textbf{3.15} & 5.61 & 50.60 & 57.80 & \textbf{13.30}	& 268.40 & 0.64 & \textbf{0.90} & 0.75 & 26.40 & 2035.66 \\
TWANG & 97.67 &	26.59 &	3.58 &	\textbf{4.62} &	\textbf{0.00} &	\textbf{6.70} &	75.80 &	193.50 &	0.77 &	\textbf{0.90}	& 0.83 &	11.02	& 4774.83 \\
TWANG+U & 97.81 & 25.30 & 3.64 & 4.75 & \textbf{0.00} & 12.40 & 103.70 & \textbf{59.50} & \textbf{0.91} & 0.89 & \textbf{0.90} & 11.15 & 4728.84 \\
\bottomrule
\vspace{1px}
\end{tabular}}
\end{center}
\captionsetup{width=1.0\hsize}
The criteria used to compare the algorithms are the Rand index (RI), the Jaccard index (JI), the normalized sum of distances (NSD) and the Hausdorff metric (HM) as described in \cite{Coelho09}. Additionally, the topological errors were assessed by counting split, merged, added and missing objects. Precision, recall and F-Score are based on the topological errors by considering split and added nuclei as false positives and merged and missing objects as false negatives, respectively. The achieved time performance was measured in seconds (smaller values are better) and voxels per second (larger values are better). All values represent the arithmetic mean of the individually processed benchmark images.
\label{tab:Segmentation:DetectionPerformance}
\end{sidewaystable}

Regarding the processing times OTSU was the fastest approach, but due to the poor quality without the uncertainty-based extension it was not really an option for a reliable analysis of the image data. TWANG and TWANG+U were $1.8$ times slower than the plain OTSU method but $2.4$ and $3.8$ times faster than OTSUWW and OTSUWW+U, respectively. At the same time, TWANG+U was the most precise approach (F-Score $0.9$). Furthermore, OTSUWW+U was $1.6$ times faster than OTSUWW due to the focused object splitting and additionally produced better results due to the improved seed detection and noise reduction.

These results confirmed that the uncertainty information could be efficiently exploited to guide computationally demanding processing operators to specific locations and thus, to speed up processing operations while the result quality was preserved or even improved.

\begin{figure*}[htbp]
\includegraphics[width=\hsize]{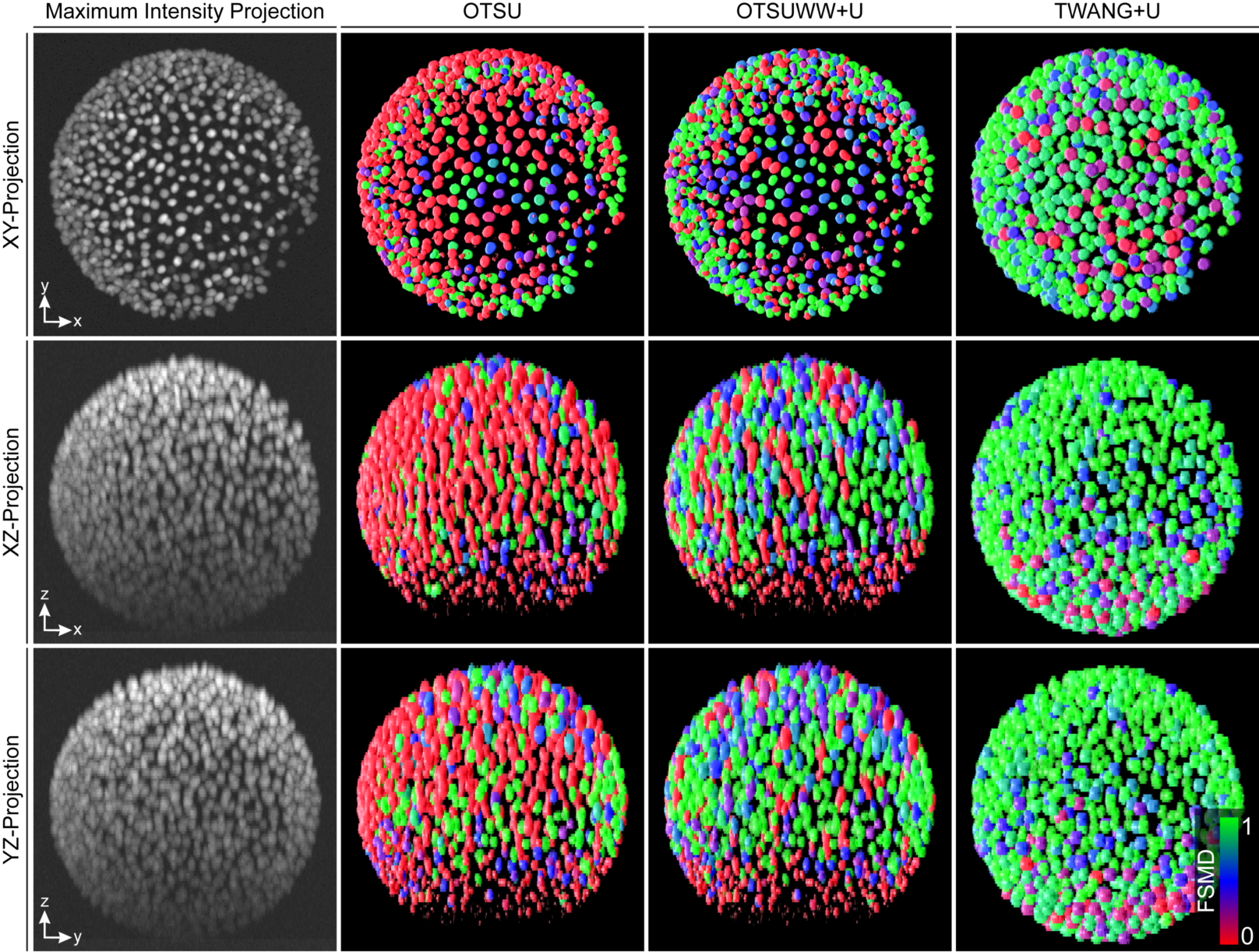}
\caption{Maximum intensity projections of the raw image and exemplary volume renderings of the automatic segmentation results produced by OTSU, OTSUWW+U and TWANG+U from different viewpoints (XY, XZ and YZ). The FSMD of individual detected objects was estimated using the morphological criteria volume and size and was used for coloring (red over blue to green for low, medium and high membership degree, respectively).}
\label{fig:Segmentation:Comparison}
\end{figure*}
Exemplary FSMDs of the final segmentation results of different algorithms are depicted in \cref{fig:Segmentation:Comparison} and provide a convenient visualization to instantly assess the segmentation quality and to identify potential problems of the methods even by non-experts. All algorithms suffered from the light attenuation in the axial direction. Especially, the techniques that relied on a single global intensity threshold had problems to identify the objects located in these low-contrast regions. Besides missing many objects, OTSU additionally merged many of the high intensity objects to a single large blob. Especially, in the z-direction many mergers occurred due to the lower sampling in this direction. However, these merged regions could be successfully split to a large extent using the proposed seed-based splitting techniques (OTSUWW, OTSUWW+U). As TWANG directly operated on the provided seeds, it was still able to extract most of the objects in these regions and yielded even higher recall values using the LoGNSM+F+U seed points. However, due to the low-contrast, the segmentation quality of the extracted segments in these regions was reduced. In \cref{tab:SeedDetection:DetectionPerformance}, this is reflected by the increased number of added objects for TWANG and TWANG+U, which were mostly no real false positives as described above.

\begin{figure}[htbp]
\includegraphics[width=\hsize]{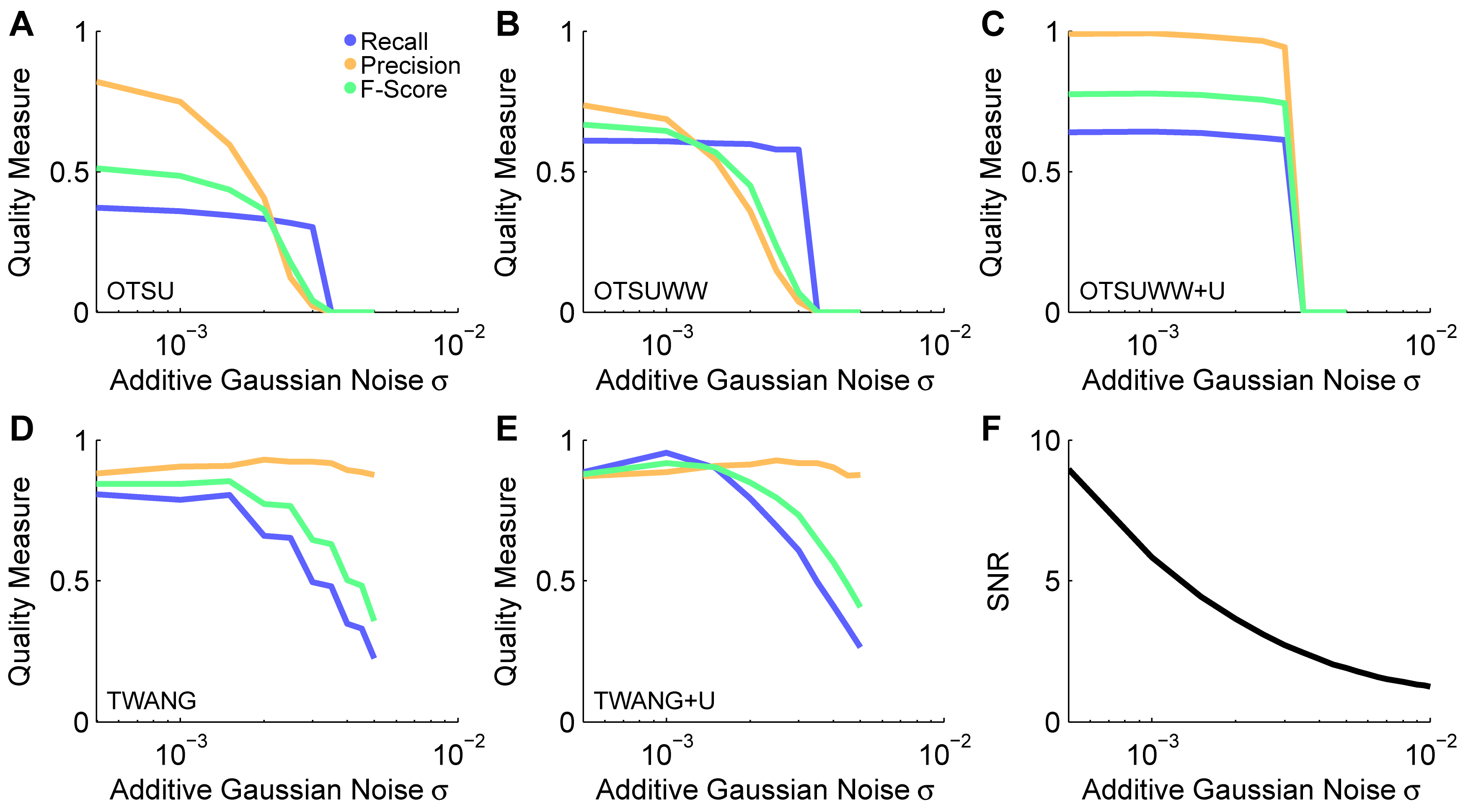}
\caption{Performance evaluation of the segmentation methods OTSU (A), OTSUWW (B), OTSUWW+U (C), TWANG (D) and TWANG+U (E) on images of the \texttt{SBDE2} data set with different signal-to-noise ratios. The methods based on adaptive thresholding (OTSU, OTSUWW) suffered from high noise levels and produced a successively increased amount of false positive detections, which could be efficiently suppressed using the uncertainty framework-based extension (OTSUWW+U). The result quality of both TWANG versions directly correlated with the quality of the provided seed points, \ie, TWANG+U benefited from the improved detection rate of LoGNSM+F+U.}
\label{fig:Segmentation:NoiseLevelInfluence}
\end{figure}

To investigate the impact of the signal-to-noise ratio of the images to the segmentation quality, the benchmark data set \texttt{SBDE2} was processed using all five algorithms. The segmentation quality of all adaptive thresholding-based methods was heavily affected by the noise level of the images yielding poor precision and recall values even for the lowest noise levels (\cref{fig:Segmentation:NoiseLevelInfluence}A, B). This was caused by the global threshold that on the one hand merged a lot of objects and on the other hand detected a high amount of false positive segments. The uncertainty-based method OTSUWW+U successfully preserved the increased recall of OTSUWW and at the same time substantially increased the precision to an almost perfect level for noise parameters of $\sigma_\text{agn} < 0.003$ (\cref{fig:Segmentation:NoiseLevelInfluence}C). The increasing number of small segments observed for OTSU and OTSUWW could be efficiently filtered using the uncertainty-based object rejection. As TWANG heavily depended on the quality of the provided seeds, the observed curves in \cref{fig:Segmentation:NoiseLevelInfluence}C, D show a high correlation to the seed detection performance (\cref{fig:SeedDetection:NoiseLevelInfluence}A, C) and render it as a suitable method even for higher noise levels. The improved seed detection of LoGNSM+F+U also directly affected the quality of the TWANG+U segmentation. Note that the seed points were adjusted for each of the noise levels, \ie, subtle variations in the precision and recall (\cref{fig:Segmentation:NoiseLevelInfluence}D, E) values are caused by a subjective manual threshold adaption.

\subsection{Tracking Validation}
\label{sec:TrackingValidation}
\begin{figure}[htbp]
\includegraphics[width=\hsize]{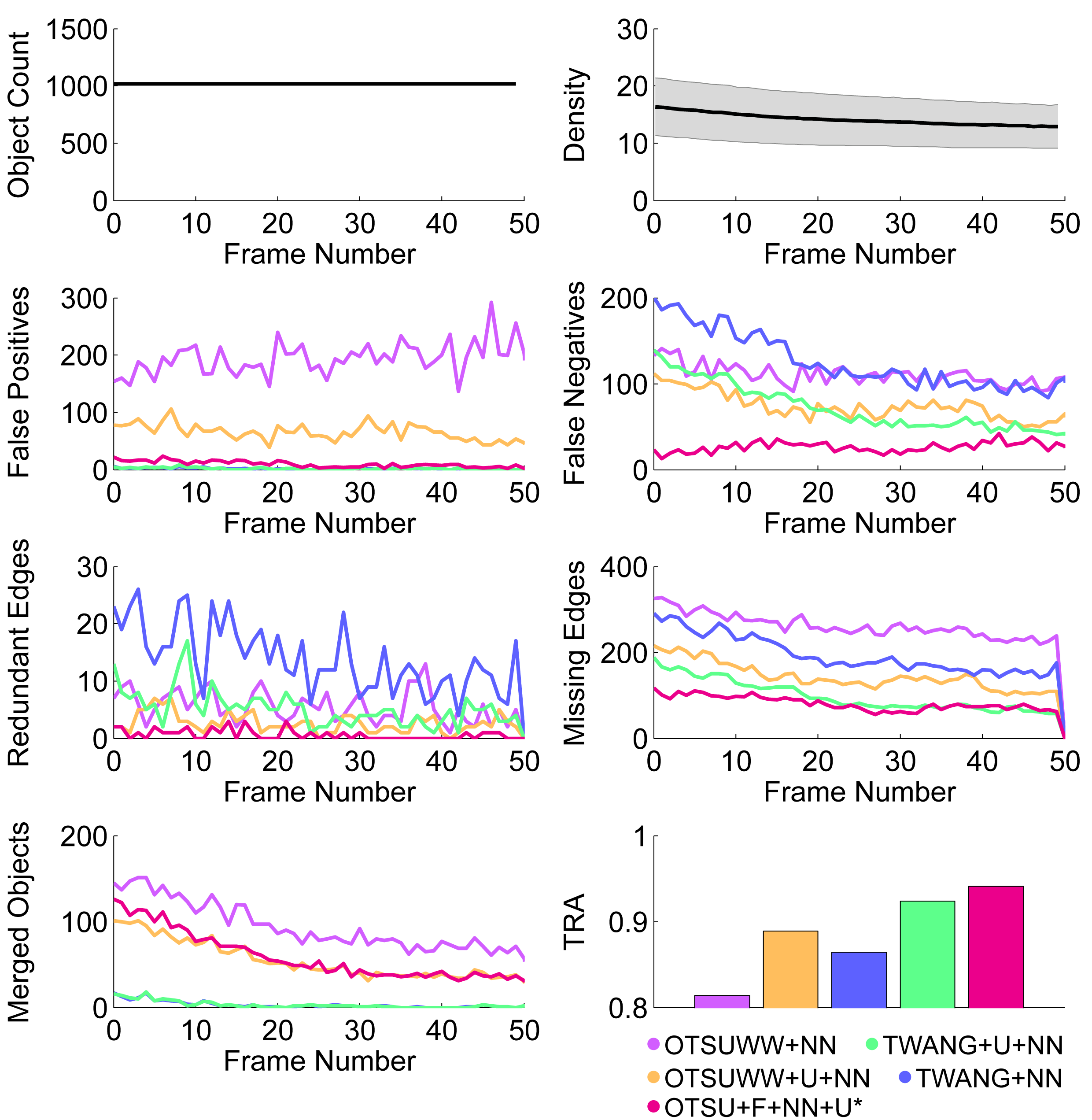}
\caption{Quantitative performance assessment of a nearest neighbor tracking algorithm (NN) applied on different segmentation results obtained on the \texttt{SBDE3} data set. Two algorithms without uncertainty-based improvements as described in \cite{Stegmaier14} (OTSUWW, TWANG) were compared to enhanced pipelines that explicitly incorporate prior knowledge-based uncertainty treatment (OTSUWW+U, TWANG+U, OTSU+F+NN+U). (*) indicates that the respective algorithm did not produce a usable segmentation image, as the correction was solely performed at the tracking step.}
\label{fig:Tracking:Results}
\end{figure}

The tracking validation was performed on the \texttt{SBDE3} data set, which consisted of $50$ frames with two simultaneously acquired rotation images for each frame, yielding a total number of $100$ frames that needed to be processed. Segmentation was performed using OTSUWW, OTSUWW+U, TWANG and TWANG+U separately for all time points and view angles. To obtain a single set of objects for each time point, segmentation results of different view angles were fused using a segment-based fusion approach described in \cref{sec:MultiviewFusion} (Additional file 1). The centroids of all detected objects were then used to perform the nearest neighbor tracking (NN). In addition, the method described in \cref{sec:Tracking} (OTSU+F+NN+U) was applied to the test data set and the Otsu-based threshold was applied to both rotation images independently and the resulting binary images were then fused by simply using the maximum pixel value of the two images. FSMD values were then estimated on the connected components of the fused image using the same fuzzy sets as for the segmentation step. The obtained tracking results are summarized in \cref{tab:Tracking:PerformanceTable} and \cref{fig:Tracking:Results}. 
\begin{sidewaystable}[htbp]
\caption{Quantitative assessment of the tracking performance}
\begin{center}
\resizebox{\hsize}{!}{
\begin{tabular}{lccccccccccccc} 
\toprule
\textbf{Method} & \textbf{TP} & \textbf{FP} & \textbf{FN} & \textbf{Redundant} & \textbf{Missing} & \textbf{Merged} & \textbf{Rec.} & \textbf{Prec.} & \textbf{F-Score} & \textbf{TRA} & \textbf{Time (s)} & \textbf{KVoxel/s} \\
\midrule
OTSUWW+NN & 905.33 & 194.10 & 110.67 & 5.51 & 256.88 & 94.33 & 0.89 & 0.82 & 0.86 & 0.81 & 51.36 & 1020.81 \\
OTSUWW+U+NN & 941.94 & 65.65 & 74.06 & 2.63 & 141.39 & 55.04 & 0.93 & 0.93 & 0.93 & 0.89 & 33.68 & 1556.68 \\
TWANG+NN & 889.73 & \textbf{1.73} & 126.27 & 13.75 & 191.16 & 3.29 & 0.88 & \textbf{1.00} & 0.93 & 0.86 & \textbf{14.27} & \textbf{3674.06} \\
TWANG+U+NN & 943.90 & 1.76 & 72.10 & 5.16 & 96.55 & \textbf{3.27} & 0.93 & \textbf{1.00} & 0.96 & 0.92 & 21.55 & 2432.89 \\
OTSU+F+NN+U$^\ast$ & \textbf{989.63} & 9.84 & \textbf{26.37} & \textbf{0.67} & \textbf{79.67} & 59.33 & \textbf{0.97} & 0.99 & \textbf{0.98} & \textbf{0.94} & 18.66 & 2809.72 \\
\bottomrule
\vspace{1px}
\end{tabular}}
\end{center}
Quantitative performance assessment of a nearest neighbor tracking algorithm (NN) applied on different segmentation results. Two algorithms without uncertainty-based improvements (OTSUWW, TWANG) were compared to enhanced pipelines that explicitly incorporated prior knowledge-based uncertainty treatment (OTSUWW+U, TWANG+U). Furthermore, an OTSU-based segmentation with additional seed points from LoGNSM+F+U (OTSU+F+NN+U) was used with an adapted tracking algorithm. Note that the segmentation produced by OTSU+F+NN+U was not usable for other purposes than tracking due to many merged regions (indicated by ($^\ast$)). The validation measures correspond to true positives (TP), false positives (FP), false negatives (FN), redundant edges (Red.), missing edges (Miss.) and merged objects (Merg.). Furthermore, recall, precision, F-Score and the TRA measure were calculated as described in \cref{sec:PerformanceAssessment} and \cite{Maska14}. Processing times are average values for applying segmentation and tracking on a single image and were measured in seconds (lower values are better) and voxels per second (higher values are better).
\label{tab:Tracking:PerformanceTable}
\end{sidewaystable}

Both pipelines without uncertainty treatment reached the lowest tracking accuracy with respect to the tracking (TRA, see \cref{sec:PerformanceAssessment}) quality measure due to an increased number of missing objects (recall of $0.89$ for OTSUWW+NN, and $0.88$ for TWANG+NN) and a high number of false positive detections (precision of $0.82$ for OTSUWW+NN). Of course, these missing objects directly correlated with the number of missing edges and explain the $222.4\%$ (OTSUWW+NN) and $139.9\%$ (TWANG+NN) higher amount of missing edges compared to the best scoring algorithm in this category (OTSU+F+NN+U). Furthermore, OTSUWW+NN suffered from many merged regions that also contributed to the $8.2\%$ lower recall value compared to OTSU+F+NN+U. In contrast to this, all uncertainty-enhanced methods provided comparable results, with the best results achieved by OTSU+F+NN+U and TWANG+U+NN. Although the amount of false negatives of TWANG+U+NN was almost halved compared to TWANG+NN, the detection rate was still the largest problem of this pipeline resulting in a $21.2\%$ higher amount of missing edges compared to OTSU+F+NN+U. On the other hand, OTSU+F+NN+U still suffered from the same under-segmentation tendency as observed for OTSUWW+NN and OTSUWW+U+NN and consequently missing edges were its main problem. With respect to false positive detections both TWANG-based methods provided the best results, with precision values of $1.0$ due to a high quality seed detection that benefitted from the simultaneous multiview acquisition. This reflects an improvement of the precision obtained by the TWANG-based methods of $22.0\%$ compared to OTSUWW+NN, which did not have an uncertainty-based object exclusion and thus an increased amount of false positive detections in background regions. The reason for the slightly higher number of redundant edges observed for the two TWANG-based methods is not yet fully clear. Most likely, the seed detection already provided a redundant seed to the segmentation method, which produced two nearby segments that were in turn counted as a redundant segment by the tracking evaluation. However, even for the worst algorithm in this category (TWANG+NN) redundant edges were only observed for $1.4\%$ of the tracked objects and thus play a minor role compared to the other tracking errors. As shown in \cref{fig:Tracking:Results}, the number of false negatives and merged objects directly correlated with the density of the objects, \ie, the closer the objects are to each other, the more under-segmentation errors occurred. However, this was not the case for TWANG-based methods due to the explicit prior knowledge about the object size that is incorporated to the algorithm. The OTSU+F+NN+U method produced the best tracking result and was the second fastest method (TRA value of $0.94$ and average processing time of $18.7s$) closely followed by TWANG+U+NN (TRA value of $0.92$ and average processing time of $21.6s$). Compared to OTSUWW+NN, TWANG+U+NN and OTSU+F+NN+U provide superior quality in all categories and in particular an increase of the TRA measure by $ 13.6\%$ and $16.0\%$, and a decrease of processing times by $58.0\%$ and $63.7\%$, respectively. Thus, the two latter methods  represent the best quality vs.\ speed trade-off and are suitable for large-scale analyses. Although OTSU+F+NN+U provided excellent results in this comparison, it should be noted that the extracted segmentation masks were largely merged and an object splitting approach as performed for OTSUWW would be required if object properties need to be known. An additional object splitting approach, however, would eradicate the performance benefit of the method and, \eg, the TWANG+U+NN pipeline should be favored in this case.

To fall back on seed point information has no benefit for segmentation methods like TWANG, where the algorithmic design already only extracts a single segment per seed point and literally no merged objects exist. However, an interesting extension to consider in upcoming work might be a combination of the LoGNSM+F+U and the OTSU-based segmentation for seed detection and to feed these seeds to the TWANG algorithm, to reach both a further reduced amount of missed objects and a reduced amount of merged objects. Moreover, the temporal coherence was not yet considered in the investigated framework, \ie, additionally allowing a nearest neighbor matching over multiple frames could potentially also help to reduce the number of missing and redundant detections.

\subsection{Application to Light-Sheet Microscopy Images of Zebrafish Embryos}
\label{sec:ApplicationExample}
\begin{figure*}[htbp]
\includegraphics[width=\hsize]{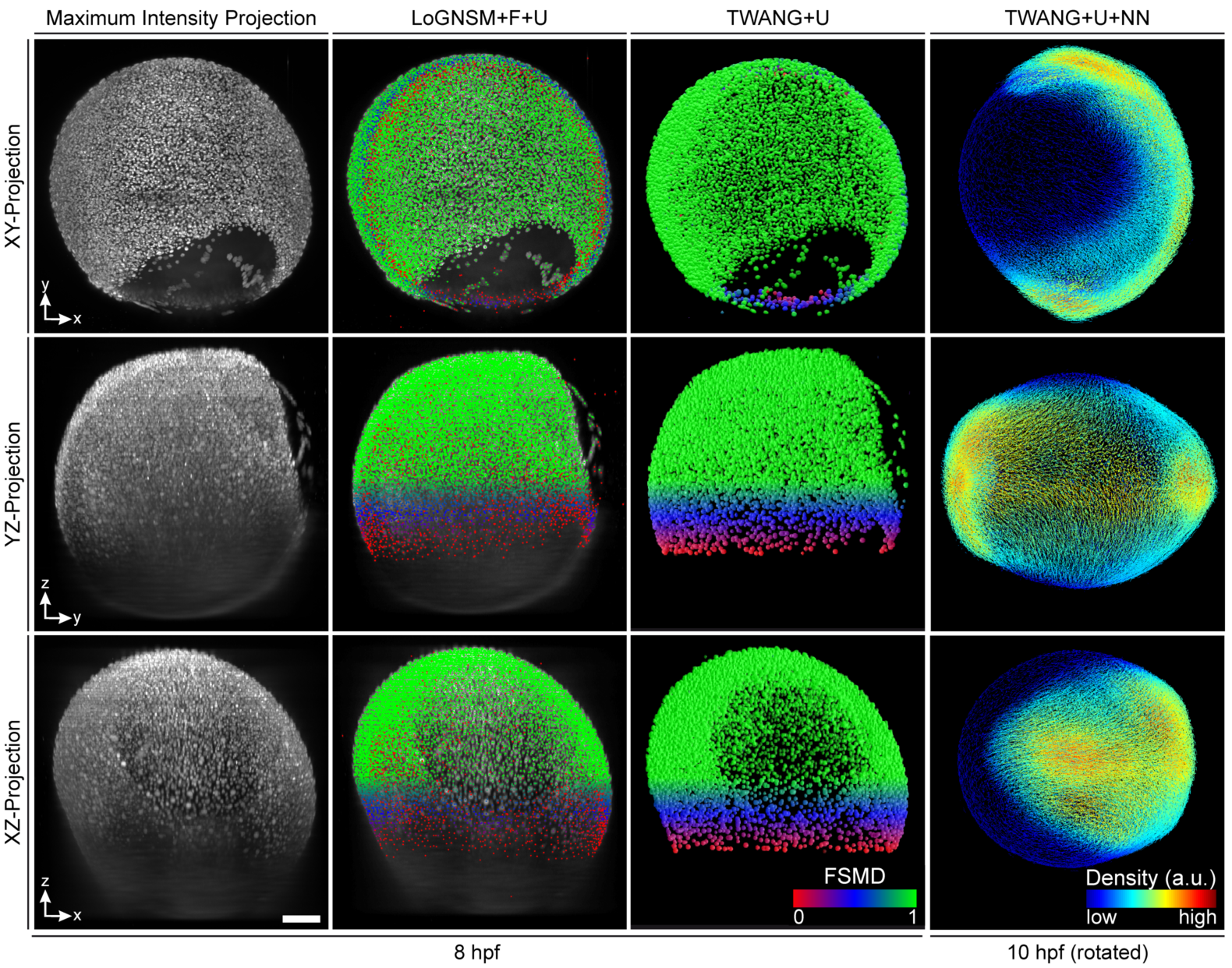}
\caption{Application of the presented framework to 3D+t light-sheet microscopy images of zebrafish embryos. The panels show the maximum intensity projections on the XY, YZ and XZ planes. Seed points were extracted using the LoGNSM+F+U method (\cref{sec:SeedDetection}) and the improved seed points were used for segmentation using the TWANG+U method (\cref{sec:Segmentation}). Extracted objects of two opposite views were combined using a segment-based fusion approach (\cref{sec:MultiviewFusion}) and the fused object locations were then used for tracking (TWANG+U+NN, \cref{sec:Tracking}). The first three columns show results of a single view at approximately 8 hours post fertilization (hpf) with the color-code according to the FSMD values of the extracted objects (red over blue to green for low, medium and high FSMD values). The last column shows the results after multiview fusion and tracking at about 10~hpf using a cell density for coloring. In addition, the embryo was oriented along the coordinate axes, such that the anteroposterior axis was aligned with the Y axis (animal pole on the top) and the dorsoventral axis aligned with the X axis (dorsal to the right) \cite{Kobitski15}. Data were taken from our previously published work \cite{Stegmaier14, Kobitski15, Stegmaier16a, Stegmaier16AT}. Scale bar, 100~$\mu$m.}
\label{fig:ApplicationExample}
\end{figure*}

The presented framework was successfully used for the automated analysis of large-scale 3D+t microscopy images of developing zebrafish embryos \cite{Stegmaier14, Kobitski15, Stegmaier16a, Stegmaier16AT}. In particular, we used the LoGNSM+U method, \ie, seed points were detected using a non-strict local maximum detection with a subsequent fusion of redundant detections and a false positive suppression based on the axial location of the seeds as well as their fluorescence intensity information (\cref{sec:SeedDetection}). These seeds were then provided to the TWANG algorithm as described in \cite{Stegmaier14} and the segments of different views were combined using a segment-based fusion approach (\cref{sec:MultiviewFusion} in Additional file 1 and \cite{Kobitski15}). Finally, a nearest-neighbor tracking was applied to the detected objects to obtain the movement trajectories (TWANG+U+NN). As there were about $8$ million dynamically interacting objects in total for the investigated time period of about 3-10 hours post fertilization (hpf), manual labeling and validation were impossible. Qualitative results of the seed detection, the segmentation and the tracking stage are depicted in \cref{fig:ApplicationExample}.

\section{Conclusion}
In this contribution, we presented a general concept for the mathematical formulation of prior knowledge and showed how image analysis pipelines can be equipped with the formalized prior knowledge to make more elaborate decisions. The framework includes the propagation of estimated result uncertainties, to be able to inform downstream pipeline operators about the validity of their input data and to potentially improve their results. Besides these general concepts, we demonstrated how an exemplary pipeline consisting of seed point detection, segmentation, multiview fusion and tracking could be systematically extended by the proposed uncertainty considerations, in order to filter, repair and fuse produced data. The performance of all proposed improvements was quantitatively assessed on a new and comprehensive validation benchmark inspired by light-sheet microscopy recordings of live specimen. The extensions proofed their superior performance compared to the plain pipelines and only had a low impact on the processing times due to the lightweight adaptions of the propagated feature matrices. Thus, the proposed framework represents a powerful approach to improve the quality and efficiency of image analysis pipelines. Several components of the presented framework were successfully used to analyze large-scale 3D+t light-sheet microscopy images of developing zebrafish embryos as described in \cref{sec:ApplicationExample} and \cite{Stegmaier12, Kobitski15}.

For simplicity, we used mostly focused on simple processing methods to illustrate the general concepts. However, extending more complex seed detection, segmentation or tracking algorithms with the presented concepts of filtering, splitting and fusion should work analogously if the uncertainty-based corrections are considered as a post-processing strategy of each processing operator. Especially the tracking step offers a lot of potential to be further improved. The uncertainty framework could be exploited to classify movement events, \eg, to detect object divisions or to reconstruct missing objects using the temporal coherence of the objects. However, this was beyond the scope of this paper and will be addressed in upcoming work. In addition to the algorithmic improvements, we showed how the respective fuzzy sets can be parameterized based on available prior knowledge, such as feature histograms or knowledge about acquisition deficiencies. The respective shape of the fuzzy sets has to be determined based on the desired outcome. For instance, the false positive suppression at the seed point detection stage could be performed with a fixed threshold instead of an explicit usage of fuzzy sets for the intensity-based features. However, to model the increasing uncertainty (decreasing FSMD value) in regions farther away from the detection objective using a trapezoidal shape was more appropriate. Further work has to be put on the automatic determination of the involved fuzzy sets, \eg, using a semi-automatic approach for a manual classification of a representative subset of data.

\section*{Acknowledgements}
The work was funded by the German Research Foundation DFG (JS, Grant No MI 1315/4, associated with SPP 1736 Algorithms for Big Data) and to the Helmholtz Association in the program BioInterfaces in Technology and Medicine (RM).


\appendix

\section{Validation Benchmark}
\label{sec:chap4:Benchmark}
To provide a thorough validation of the entire pipeline comprised of seed point detection, segmentation, multiview fusion and tracking, we used our recently presented approach for generating comprehensive validation benchmarks \cite{Stegmaier16a} with an adapted object movement simulation. In brief, object locations, object movements and object interactions were simulated over multiple time points to obtain movement behaviors that resembled biological specimens. At each simulated position, a video snippet containing a simulated fluorescent cellular nucleus \cite{Svoboda12} was added to an artificial 3D image, to form the data basis for each  time point. To simulate acquisition deficiencies, the simulated images were disrupted by additive Gaussian noise, Poisson shot noise, a point-spread-function simulation \cite{Preibisch14}, light attenuation and multiview acquisition simulation as detailed in \cite{Stegmaier16a}. All acquisition deficiencies can be added to the simulated raw images using an XPIWIT pipeline \cite{Bartschat16} and the pipeline can be downloaded from \url{https://bitbucket.org/jstegmaier/embryomicsbenchmark/downloads/AcquisitionSimulationPipeline.zip}.

\begin{figure*}[htbp]
\centerline{\includegraphics[width=\textwidth]{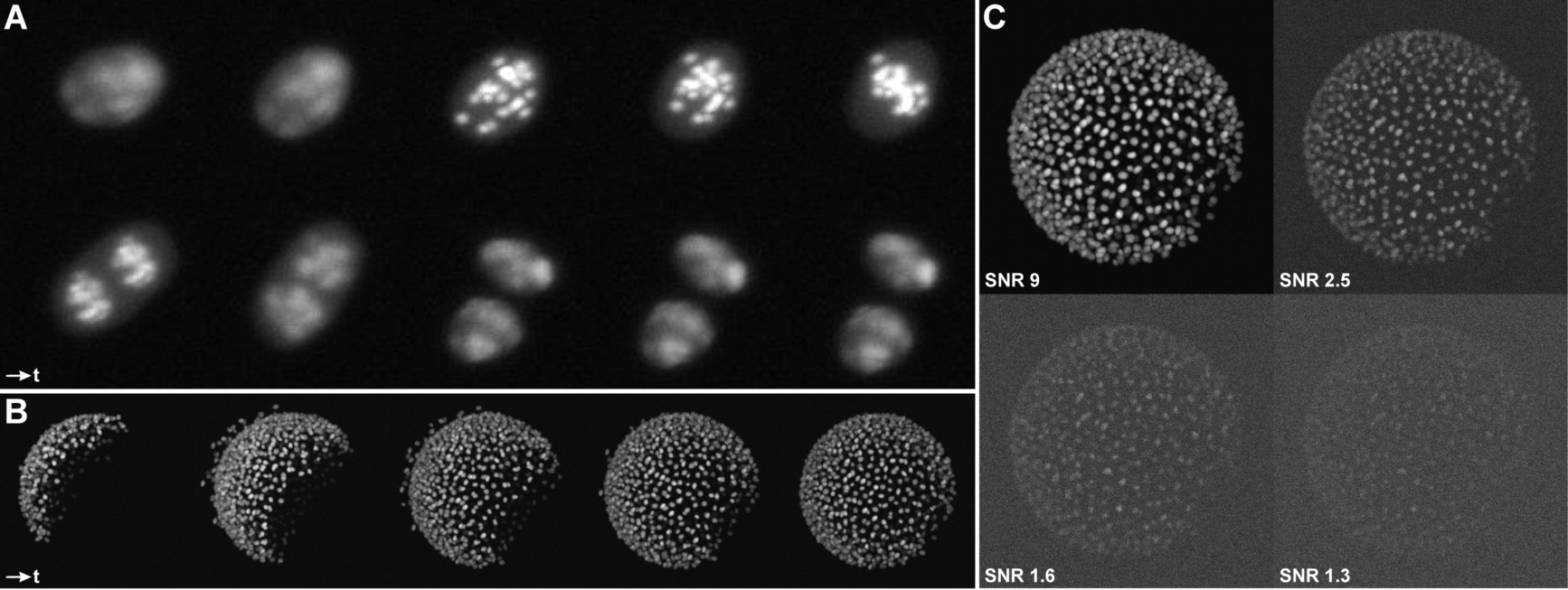}}
\caption{Maximum intensity projections of simulated benchmark images. (A) Maximum projections of an extracted division cycle of one simulated nucleus \cite{Svoboda12}. Time increases from left to right and top to bottom. Single objects were randomly initialized and simulated for a predefined experimental duration. This approach yielded a simulated embryo including object movement, object interaction and object divisions with available ground truth (B). Generated raw image sequences were manipulated to simulate various acquisition conditions, such as different levels of additive Gaussian noise (C). The ground truth enabled a quantitative analysis of the algorithmic performance on realistic image data.}
\label{fig:chap4:Benchmark:TimeSeries}
\end{figure*}

For the sake of simplicity, we simulated objects that were moving on a spherical surface. Instead of using object displacements of a real embryo as described in \cite{Stegmaier16a}, the simulated objects only moved due to density changes and the resulting repulsive and adhesive forces acting between neighboring objects. Furthermore, the simulation was constrained to a spherical surface to prevent arbitrary movement in the simulation space. A schematic illustration of the simulated specimen and an overview of the involved simulation steps of the benchmark is shown in \cref{fig:chap4:Benchmark:Illustration}. The repulsive ($\Delta\mathbf{x}^{\text{rep}}$) and adhesive forces ($\Delta\mathbf{x}^{\text{adh}}$) as well as the parameterization was taken from \cite{Macklin12} and the boundary constraint was defined as:
\begin{gather}
	\Delta\mathbf{x}^{\text{bdr}}(\mathbf{x}, \mathbf{c}, r_i, r_o, a) = 
	\begin{cases} 
		\frac{\mathbf{x}-\mathbf{c}}{\Vert\mathbf{x}-\mathbf{c}\Vert} \cdot \left( 1-\frac{1}{e^{-a(\Vert\mathbf{x}-\mathbf{c}\Vert-r_i)}} \right), & \Vert\mathbf{x}-\mathbf{c}\Vert < r_i \\ 
		-\frac{\mathbf{x}-\mathbf{c}}{\Vert\mathbf{x}-\mathbf{c}\Vert} \cdot \left(1-\frac{1}{e^{a(\Vert\mathbf{x}-\mathbf{c}\Vert-r_o)}} \right), & \Vert\mathbf{x}-\mathbf{c}\Vert > r_o \\ 
		\mathbf{0}, & \text{else}.
	 \end{cases} 
\label{eq:chap4:Benchmark:BoundaryForce}
\end{gather}
In \cref{eq:chap4:Benchmark:BoundaryForce}, $\mathbf{x}$ is the centroid of the considered object, $\mathbf{c}$ is the center of the bounding volumes, $r_i$ and $r_o$ are the radii of the inner and the outer sphere, respectively, and finally $a$ controls the shape of the sigmoidal boundary potential function. $\Delta\mathbf{x}^{\text{bdr}}$ only contributed to the displacement of a simulated object if the object was already out of the boundary. This additional displacement component prevented objects from entering the inner bounding sphere and from escaping the outer bounding sphere of the simulated embryo (\cref{fig:chap4:Benchmark:Illustration}A). Analogous to the formulation used in \cite{Stegmaier16a}, the total displacement vector of a single object at a given time point can be summarized to:
\begin{figure*}[htbp]
\centerline{\includegraphics[width=\hsize]{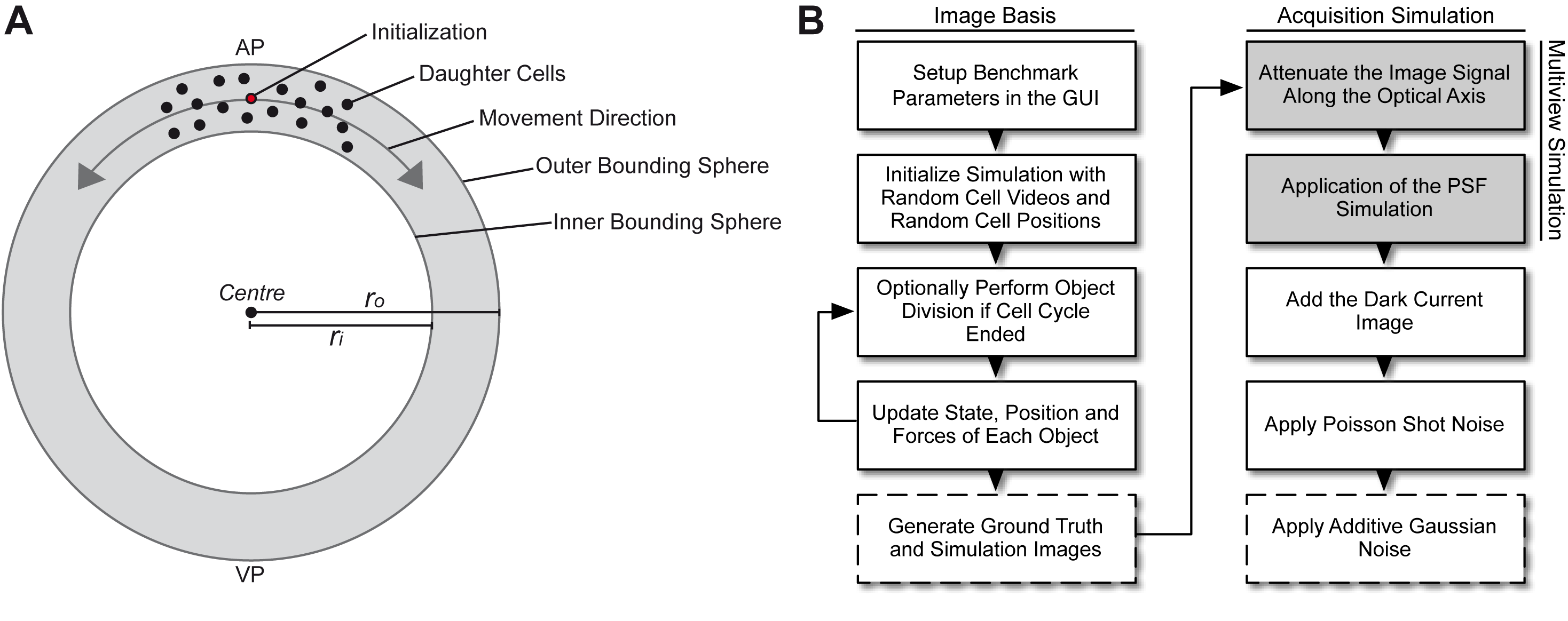}}
\caption{Pipeline schematic of the benchmark simulation. (A) Illustration of the embryo simulation. Starting with an initial object, multiple division cycles were simulated including object interaction, object divisions and morphological constraints. Due to inner and outer bounding spheres and the between-object interactions, cells migrated from the animal pole (AP) to the artificial vegetal pole (VP), which produced a behavior similar to the epiboly movement of zebrafish embryogenesis. (B) The performed steps for a realistic simulated 3D+t benchmark. The left column reflects the object simulation and returned raw images that contained dynamic objects and the associated ground truth data. The right column contains the acquisition simulation, which distorted the simulated images by an artificial signal attenuation, a point spread function simulation (PSF), a dark current image simulation, Poisson distributed photon shot noise and additive Gaussian noise. Steps shaded in gray could optionally consider image rotation, if a multiview experiment was simulated and the output operators are indicated by dashed edge lines (adapted from \cite{Stegmaier16a}).}
\label{fig:chap4:Benchmark:Illustration}
\end{figure*}

\begin{align}
	\Delta\mathbf{x}_i^{\text{tot}} = &w_{\text{bdr}} \cdot \Delta\mathbf{x}^{\text{bdr}}(\mathbf{x}_i) + \nonumber \\
	 & \sum_{j\in\{1,...,N\}}^{i\neq j}{\left[w_{\text{rep}} \cdot \Delta\mathbf{x}^{\text{rep}}(\Vert\mathbf{x}_i-\mathbf{x}_j\Vert) + w_{\text{adh}} \cdot \Delta\mathbf{x}^{\text{adh}}(\Vert\mathbf{x}_i-\mathbf{x}_j\Vert)\right]}.
	\label{eq:chap4:Benchmark:TotalForce}
\end{align}
The weights of the adhesive and repulsive displacement components were set to the default values mentioned in \cite{Macklin12}: $w_{\text{adh}}=0.52$, $w_{\text{rep}}=1.0$. Moreover, the weight for the boundary constraint $w_{\text{bdr}}=3.0$ was manually adjusted, such that the interacting objects remained within the spherical boundaries. Note that these parameters were empirically determined and that the presented model does not necessarily represent an accurate physical simulation of the interacting objects. However, the determined parameters produced movement behaviors that were similar to the epiboly movements observed during early zebrafish development due to increased object densities and boundary constraints that caused a directed movement on a sphere surface. Exemplary benchmark images are shown in \cref{fig:chap4:Benchmark:TimeSeries} for different time points and different additive Gaussian noise levels. Due to the availability of a complete ground truth, a detailed quantitative analysis of all involved pipeline steps could be performed with a single benchmark using the performance measures described in the following section. All validation datasets ($\approx$70GB) used in this publication are available upon request.

\section{Performance Assessment}
\label{sec:PerformanceAssessment}
\subsection{Seed Detection} 
The seed detection quality was evaluated using the benchmark datasets \texttt{SBDE1} and \texttt{SBDE2} (\cref{tab:Appendix:BenchmarkDatasetsEmbryo}). The intersections of the detected seeds with the labeled ground truth image were calculated. True positives (TP) were counted as ground truth objects that contained at least one seed point. Seed points that were detected in background regions or redundant detections of ground truth objects were considered as false positives (FP). Ground truth objects that did not contain a seed point were counted as false negatives (FN). Using TP, FP and FN, recall, precision and the F-Score (harmonic mean of precision and recall) were calculated. For all true positives, the average distance to the centroids of the respective ground truth objects was additionally calculated.

\subsection{Segmentation}
The segmentation quality was assessed using the benchmark datasets \texttt{SBDE1} and \texttt{SBDE2} (\cref{tab:Appendix:BenchmarkDatasetsEmbryo}). As the provided ground truth of the benchmark contained the complete label images of each frame, a detailed quantitative assessment of the automatic segmentation quality could be performed. The set of segmentation validation measures proposed by Coelho \etal\ was used, namely the Rand index (RI), the Jaccard index (JI), the normalized sum of distances (NSD) and the Hausdorff metric (HM). A detailed description of the measures can be found in \cite{Coelho09}. Topological errors produced by the automatic segmentation were separated into added, missing, split or merged objects. Besides the error counts, this topological information was used to define the number of false positives as the sum of split and added cells and analogously the false negatives as the sum of merged and missing cells. These values were then used to calculate recall, precision and F-Score.

\subsection{Tracking}
To assess the tracking quality, the \texttt{SBDE3} dataset was used (\cref{tab:Appendix:BenchmarkDatasetsEmbryo}). The comparison of the investigated algorithms was performed using the \texttt{TRA} measure as described by Ma{\v{s}}ka \etal \cite{Maska14}. This measure was calculated by considering the tracking result as an acyclic oriented graph and by comparing this graph to the respective ground truth graph. The inverted, weighted and normalized number of required changes to transform the automatically generated graph to the ground truth graph yielded the normalized \texttt{TRA} measure (higher values are better with 1 being ideal). As the centroids of all objects, the complete temporal association and the object ancestry was known for the simulated \texttt{SBDE3} dataset, the required ground truth graph could directly be generated using this data. To obtain a more detailed view on the errors made by the respective tracking algorithms, the number of false positive detections, false negative detections, incorrect edges, missing edges, redundant edges and merged objects were counted. Detailed descriptions of the validation measures are provided in \cite{Maska14}.

\subsection{Evaluation Platform}
\label{sec:Appendix:EvaluationPlatform}
All measurements with respect to processing times were performed on a desktop PC equipped with an Intel Core i7-2600 CPU @ 3.4GHz and 32GB of memory installed using the Windows 7 (x64) operating system.

\section{Multiview Fusion}
\label{sec:MultiviewFusion}
A frequently used approach to partly overcome quality deficiencies in 3D microscopy that are caused by attenuation and scattering of light along the axial direction is the acquisition of multiple views from different perspectives \cite{Krzic12, Chhetri15}. Such complementary image stacks of a specimen can be obtained either by using multiple oppositely arranged detection paths \cite{Tomer12, Chhetri15} or by a rotation of the probe using a single camera \cite{Kobitski15}. Both strategies for the acquisition of multiview images require a subsequent fusion step of the information present in the individual views into a single consistent representation. The benchmark dataset \texttt{SBDE3} used for the tracking validation contains such a simulated multiview acquisition, \ie, at each time point, two opposite images were simulated. To fuse to complementary views, we use a segmentation-based fusion approach that combines the results of separately applied segmentation on the different views to a single segmentation image. As the image transformation of $180^{\circ}$ was already known from the benchmark generation step, the registration process was skipped. Corresponding segments were identified using a histogram-based approach, \ie, a 2D label histogram was filled by iterating over all voxels of both 3D segmentation images and by successively increasing the 2D histogram bins indicated by the respective image label pairs. Segments that were only present in one or the other image could be found by searching for empty columns and rows of the histogram, respectively, and by copying these segments without any further consideration into the new result image. In the next step, the assignments of the remaining segments in both images were identified by searching for the respective label pair with the maximum overlap. In the case of segmentation fusion, it was not desired to perform a weighted average of the segments but rather to use the better segment for the new image. This was accomplished by simply selecting the segment with a higher FSMD value to the desired class of objects based on the FSMD values of the segments and the seed points that contained the information about the validity of the segment and the axial localization of the extracted objects, respectively. The final segmentation image was then used for tracking.

\section{Availability of Data and Material}
All validation data sets used in this publication ($\approx$70GB) are made available upon request. MATLAB scripts for generating benchmark data sets of simulated embryos as well as the XPIWIT pipeline that was used for acquisition simulation can be found at \url{https://bitbucket.org/jstegmaier/embryomicsbenchmark/}. Furthermore, source code as well as binary packages of XPIWIT are available from \url{https://bitbucket.org/jstegmaier/xpiwit/}.

\begin{table*}[htbp]
\begin{center}
\caption{The benchmark datasets used for the validation experiments. Columns list the dataset identifier (Name), the number of images (Im.), the number of noise levels (N.), the number of objects (Objects), the image resolution, the noise parameter ranges for $\sigma_{\text{agn}}$ and the dataset description.}
\resizebox{\textwidth}{!}{
\small\begin{tabular}{lcccccm{0.28\columnwidth}}
\toprule
\textbf{Name} & \textbf{Im.} & \textbf{N.} & \textbf{Objects} & \textbf{Resolution} & $\sigma_{\text{agn}}$ & \textbf{Description} \\
\midrule
\texttt{SBDE1} & 10 & 1 & 300--1000 & 640$\times$640$\times$128 & 0.0007 & Five pairs of subsequent frames with different numbers of objects. Used for seed detection and segmentation validation. \\
\texttt{SBDE2} & 20 & 20 & 1000 & 640$\times$640$\times$128 & $[0.0005,0.01]$ &  Single frame of \texttt{SBDE1} with 20 different noise levels. Used for seed detection and segmentation validation. \\
\texttt{SBDE3} & 100 & 1 & 1000 & 640$\times$640$\times$128 & 0.0007 &  Sequential time points containing $1000$ objects each. Images are rotated by $180^\circ$ at every frame to simulate a simultaneous acquisition multiview experiment SiMV ($2 \times 50$ images). Used for the validation of the tracking. \\
\bottomrule
\end{tabular}}
\label{tab:Appendix:BenchmarkDatasetsEmbryo}
\end{center}
\end{table*}

\begin{table*}[htbp]
\caption{Abbreviations, parameterizations and descriptions of the investigated seed detection algorithms.}
\resizebox{\textwidth}{!}{
\begin{tabular}{m{0.2\columnwidth}m{0.3\columnwidth}m{0.41\columnwidth}}
\toprule
\textbf{Method}& \textbf{Parameters}& \textbf{Description} \\
\midrule
LoGSM	& $\sigma_{\text{min}}=4$, $\sigma_{\text{max}}=8$, $\sigma_{\text{step}}=1$, $t_{\text{wmi}}=0.0025$ & Seed detection in the LoG scale space maximum projection with a manually adjusted window mean intensity threshold ($t_{\text{wmi}}$) and a strict maximum detection. \\
\midrule
LoGNSM & $\sigma_{\text{min}}=4$, $\sigma_{\text{max}}=8$, $\sigma_{\text{step}}=1$, $t_{\text{wmi}}=0.0025$ & Seed detection in the LoG scale space maximum projection with a manually adjusted window mean intensity threshold ($t_{\text{wmi}}$) and a non-strict maximum detection. \\
\midrule
LoGNSM+F & $\sigma_{\text{min}}=4$, $\sigma_{\text{max}}=8$, $\sigma_{\text{step}}=1$, $t_{\text{wmi}}=0.0025$, $t_{\text{dbc}}=5$ & Same detection as LoGNSM but with additional fusion (F) of redundant detections using a hierarchical clustering approach with a distance-based cutoff value ($t_{\text{dbc}}$). \\
\midrule
LoGNSM+F+U & $\sigma_{\text{min}}=4$, $\sigma_{\text{max}}=8$, $\sigma_{\text{step}}=1$, $t_{\text{dbc}}=5$, $\pmb{\theta}_{\text{wmi}} = (0.0025, 0.0025, \infty, \infty)$, $\pmb{\theta}_{\text{smi}} = (0.0007, 0.0007, \infty, \infty)$, $\pmb{\theta}_{\text{zpos}} = (50, 250, \infty, \infty)$, $\alpha_{11} = 0.0001$, $\beta_{11}=\alpha_{11}$ & Same detection as LoGNSM but with uncertainty-based (U) threshold and the fusion of LoGNSM+F. The forward threshold $\alpha_{11}$ is set slightly above zero, such that obvious false positives are rejected. As no further processing was needed $\beta_{11}$ was set to $\alpha_{11}$. \\
\bottomrule
\end{tabular}}
\label{tab:chap4:SeedDetection:ParameterizationTable}
\end{table*}

\newpage
\begin{table}[htbp]
\caption{Abbreviations, parameters and descriptions of the segmentation algorithms.}
\resizebox{\textwidth}{!}{
\begin{tabular}{m{0.18\textwidth}m{0.14\textwidth}m{0.60\textwidth}}
\toprule
\textbf{Method}& \textbf{Parameters}& \textbf{Description} \\
\midrule
OTSU	& $\sigma_{\text{smooth}}=1$ & Adaptive thresholding using Otsu's method applied on a regularized raw input image \cite{Otsu79}. \\
\midrule
OTSUWW & $\sigma_{\text{smooth}}=1$ & Same segmentation as OTSU. Merged regions in the entire image were split based on cleaned seeds of the LoGSM method, using a seeded watershed approach. \\
\midrule
OTSUWW+U & $\sigma_{\text{smooth}}=1$, $\alpha_{21}=0.1$, $\beta_{21}=0.1$ & Same segmentation as OTSU. Small noise objects with FSMD values below $\alpha_{21}=0.1$ were rejected from further processing. Merged regions with FSMD below $\beta_{21}=0.1$ that were larger than expected were split locally and in parallel using the cleaned and fused seeds of the LoGNSM+F+U method and a seeded watershed approach in small cropped regions. \\
\midrule
TWANG & $\sigma_{\text{grad}}=3$, $\sigma_{\text{kernel}}=3$, $\omega_{\text{kpm}}=1.41$ & TWANG segmentation as described in \cite{Stegmaier14} using the cleaned seeds provided by the LoGSM method. The manually optimized parameter $\omega_{\text{kpm}}=1.41$ yielded better results than the default value of $1.0$ for this dataset.  \\
\midrule
TWANG+U & $\sigma_{\text{grad}}=3$, $\sigma_{\text{kernel}}=3$, $\omega_{\text{kpm}}=1.41$, $\alpha_{21}=0$, $\beta_{21}=0$ & TWANG segmentation as described in \cite{Stegmaier14} using the cleaned and fused seeds provided by the LoGNSM+F+U method. $\alpha_{21}=0$, $\beta_{21}=0$ are adjusted such that all segments are propagated unchanged. \\
\bottomrule
\end{tabular}}
\label{tab:chap4:Segmentation:ParameterizationTable}
\end{table}


\end{document}